\documentclass[11pt]{article}
\usepackage{amssymb}
\usepackage{amsmath}
\usepackage{amsfonts}
 \usepackage{pifont}
\usepackage{amsmath,amsfonts,amssymb,theorem,graphicx,listings,txfonts}
\usepackage{graphics}
 \usepackage{caption2}
\usepackage{flafter}
 \usepackage{indentfirst}
\usepackage{epsfig}
  \usepackage{mathrsfs}
  \usepackage{float}
\usepackage{subfigure}
 \usepackage{cite}
\newtheorem{theorem}{Theorem}[section]

\newtheorem{definition}[theorem]{Definition}

\theoremstyle{example}
\newtheorem{example}[theorem]{Example}
\theoremstyle{programme}

\theoremstyle{property}

\theoremstyle{problem}

\title{Double-quantitative $\gamma^{\ast}-$fuzzy coverings approximation operators}

\author
{Guangming Lang$^{1,2,3}$ 
\thanks{Corresponding author.\quad Tel./fax: +86 021 69585800,
\newline\mbox{}\hspace{0.55cm}
E-mail address: langguangming1984@tongji.edu.cn(G.M.Lang). }\\
\small {$^{1}$ Department of Computer Science and Technology, Tongji University}\\
\small {Shanghai 201804, P.R. China}\\
\small {$^{2}$ School of Mathematics and Computer Science, Changsha University of Science and Technology}\\
\small {Changsha, Hunan 410114, P.R. China}\\
\small {$^{3}$ The Key Laboratory of Embedded System and Service Computing, Ministry of Education, Tongji University}\\
\small {Shanghai 201804, P.R. China}}
\date{}

\begin{document}
\maketitle \baselineskip=17pt
\begin{center}
\begin{quote}
{{\bf Abstract.}
In digital-based information boom, the fuzzy covering rough set model is an important mathematical tool for artificial intelligence, and how to build the bridge between the fuzzy covering rough set theory and Pawlak's model is becoming a hot research topic. In this paper, we first present the $\gamma-$fuzzy covering based probabilistic and grade approximation operators and  double-quantitative approximation operators. We also study the relationships among the three types of $\gamma-$fuzzy covering based approximation operators. Second, we propose the $\gamma^{\ast}-$fuzzy coverings based multi-granulation probabilistic and grade lower and upper approximation operators and multi-granulation double-quantitative lower and upper approximation operators. We also investigate the relationships among these types of $\gamma-$fuzzy coverings based approximation operators. Finally, we employ several examples to illustrate how to construct the lower and upper approximations of fuzzy sets with the absolute and relative quantitative information.

\\}
\end{quote}
\end{center}
\renewcommand{\thesection}{\arabic{section}}

\section{Introduction}

Rough set theory, proposed by Pawlak in 1982, is an important mathematical tool for dealing with imprecise and uncertain information in data analysis. In theoretical aspects, by extending the equivalence relation, Pawlak's model has been generalized to covering-based rough sets, fuzzy covering-based rough sets, dominance rough sets, fuzzy rough sets, rough fuzzy sets, decision-theoretic rough sets, double-quantitative rough sets, multi-granulation rough sets, and so on. In application aspects, rough set theory has been successfully applied to various fields such as machine learning, data mining, image processing, and knowledge discovery, and the applied fields are being increasing with the development of rough set theory.

Among all generalizations of Pawlak's model, fuzzy covering rough set theory computes the lower and upper approximations of fuzzy sets in fuzzy covering approximation spaces, and provides an important mathematical tool for knowledge discovery. So far, many types of fuzzy covering-based lower and upper approximation operators have been proposed with respect to different backgrounds. Especially, the fuzzy $\gamma-$covering based lower and upper approximation operators introduced by Ma\cite{Ma1,Ma2} builded the link between the fuzzy covering rough set theory and Pawlak's model, which provides an effective approach to studying the fuzzy covering approximation spaces from the view of Pawlak's rough sets. In practical situations, there are a lot of fuzzy covering approximation spaces. Especially, there exists a great number of fuzzy covering information systems. To perform knowledge discovery of fuzzy covering information systems, we should construct the effective lower and upper approximation operators for the fuzzy covering approximation spaces. There are many effective rough set models such as probabilistic rough sets, grade rough sets, double quantitative rough sets, and multi-granulation rough sets, and we should study how to construct the lower and upper approximation operators for fuzzy covering approximation spaces with the advantages of different rough set models.

The purpose of this work is shown as follows. First, we propose the fuzzy $\gamma$-covering based probabilistic lower and upper approximation operators, as extensions of probabilistic lower and upper approximation operators, in fuzzy $\gamma$-covering approximation spaces. We present the fuzzy $\gamma$-covering based grade lower and upper approximation operators as extensions of grade lower and upper approximation operators. We also discuss the relationship between the fuzzy $\gamma$-covering based probabilistic operators and the fuzzy $\gamma$-covering based grade operators. Second, we provide the fuzzy $\gamma$-covering based disjunctive double-quantitative lower and upper approximation operators in fuzzy $\gamma$-covering approximation spaces. We propose the fuzzy $\gamma$-covering based conjunctive double-quantitative lower and upper approximation operators. We also discuss the relationship between the fuzzy $\gamma$-covering based disjunctive and conjunctive double-quantitative approximation operators and the fuzzy $\gamma$-covering based probabilistic and grade approximation operators. Third, we present the fuzzy $\gamma^{\ast}$-coverings based multi-granulation probabilistic and grade lower and upper approximation operators in fuzzy $\gamma^{\ast}$-coverings information systems. We also discuss the relationship between the $\gamma^{\ast}$-coverings based multi-granulation probabilistic and grade approximation operators and the fuzzy $\gamma^{\ast}$-coverings based probabilistic and grade approximation operators. Fourth, we provide the fuzzy $\gamma^{\ast}$-coverings based disjunctive and conjunctive multi-granulation double-quantitative lower and upper approximation operators. We also discuss the relationship between the fuzzy $\gamma^{\ast}$-coverings based double-quantitative multi-granulation approximation operators and the fuzzy $\gamma^{\ast}$-coverings based multi-granulation probabilistic and grade approximation operators.

The reminder of this paper is organized as follows. Section 2 reviews the related works. In Section 3, we recall some basic concepts of covering rough set theory, probabilistic rough set theory, grade rough set theory, and fuzzy covering rough set theory. Section 4 proposes the fuzzy $\gamma-$covering based probabilistic and grade lower and upper approximation operators. In Section 5, we present the fuzzy $\gamma-$covering double-quantitative lower and upper approximation operators. Section 6 introduces the fuzzy $\gamma^{\ast}-$coverings multi-granulation lower and upper approximation operators. In Section 7, we construct the fuzzy $\gamma^{\ast}-$coverings multi-granulation double-quantitative lower and upper approximation operators. The paper ends with conclusions in Section 8.

\section{Review of related works}

In this section, we review some works related to double-quantitative rough set theory, multi-granulation rough set theory, and fuzzy covering-based rough set theory.

\indent{\bf(1) Double-quantitative rough set theory:}

Double-quantitative rough set theory\cite{Fan1,Fang1,Li4,Xu1,Zhang1,Zhang3,Zhang4,Zhang5}, as the combination of probabilistic rough sets\cite{Liang1,Liu1,Liu2,Ma2,Sang1,Sun1,Yao1,Yao2,Yao3,Yao5,Yu1} and grade rough sets\cite{Liu4,Huang1,Yao6}, considers the relative and absolute quantitative information when constructing the lower and upper approximation operators. For example, Fan et al.\cite{Fan1} proposed a couple of double-quantitative decision-theoretic rough fuzzy set models based on logical conjunction and logical disjunction operation and discuss rules and the inner relationship between two models. Fang et al.\cite{Fang1} presented the probabilistic graded rough set as an extension of Pawlak's rough set and grade rough sets and double relative quantitative decision-theoretic rough set models. Li et al.\cite{Li4} provided double-quantitative decision-theoretic rough sets and studied its properties. Xu et al.\cite{Xu1} proposed the lower and upper approximations of generalized multi-granulation double-quantitative rough sets by introducing the lower and upper support characteristic functions. They also constructed the approximation accuracy to show the advantage of the proposed model. Zhang et al.\cite{Zhang2,Zhang3} provided information architecture, granular computing and rough set model in the double-quantitative approximation space of precision and grade. They also proposed two basic double-quantitative rough set models of precision and grade and their investigation using granular computing.

\indent{\bf(2) Multi-granulation rough set theory:}

Many efforts have focused on multi-granulation rough set theory\cite{Feng1,Li2,Li3,Liang2,Lin1,Lin2,Liu3,Qian2,Qian3,Raghavan1,She1,Wu1,Xu1,Xu2,Xu3,Yang3,Zhang1}. For example, Feng et al.\cite{Feng1} proposed variable precision multi-granulation decision-theoretic
fuzzy rough sets. Li et al.\cite{Li2} presented a comparative study of multi-granulation
rough sets and concept lattices via rule acquisition. Li et al.\cite{Li3} provided multi-granulation decision-theoretic rough sets in ordered information systems. Liang et al.\cite{Liang2} established an efficient feature selection algorithm with a multi-granulation view.
Lin\cite{Lin1} introduced an approach to feature selection via neighborhood multi-granulation fusion.
Lin et al.\cite{Lin2} presented a fuzzy multi-granulation decision-theoretic
approach to multi-source fuzzy information systems.
Liu et al.\cite{Liu3} provided the multi-granulation rough sets in covering context. Qian et al.\cite{Qian2,Qian3} introduced multi-granulation rough set theory by extending Pawlak's model. Raghavan et al.\cite{Raghavan1} explored the topological properties of multi-granulation rough sets. She et al.\cite{She1} deeply studied topological structures and properties of multi-granulation rough sets. Wu et al.\cite{Wu1} proposed a formal approach to granular computing with multi-scale data decision information systems. Xu et al.\cite{Xu1,Xu2,Xu3} considered variable, fuzzy and ordered multi-granulation rough set models.  Yang et al.\cite{Yang3} updated multi-granulation rough approximations with increasing of granular structures. Zhang et al.\cite{Zhang1} provided constructive methods of rough approximation
operators and multi-granulation rough sets.

\indent{\bf(3) Fuzzy covering-based rough set theory:}

Fuzzy covering-based rough set theory\cite{Chen1,Deng1,D'eer1,Feng2,Huang1,Lang1,Li1,Ma1,Restrepo1,Seselja1,Wang1,
Yang1,Yang2,Yao4,Zhang6} has attracted more and more attention. For example, based on fuzzy covering and binary fuzzy logical operators, Deng\cite{Deng1} proposed an approach to fuzzy rough sets in the framework of lattice theory, and presented a link between the generalized fuzzy rough approximation operators and fundamental morphological operators. Feng et al.\cite{Feng2} defined a novel pair of belief and plausibility functions by employing a method of non-classical probability model and the approximation operators of a fuzzy covering. Huang et al.\cite{Huang1} presented an intuitionistic fuzzy graded covering rough set and studied its properties. Li et al.\cite{Li1} showed a general framework for the study of covering-based fuzzy approximation operators in which a fuzzy set can be approximated by some elements in a crisp or a fuzzy covering of the universe. Ma\cite{Ma1} presented the concepts of fuzzy $\gamma$-covering and fuzzy $\gamma$-neighborhood and two new types of fuzzy covering rough set models which links covering rough set theory and fuzzy rough set theory. $\check{S}e\check{s}elja$\cite{Seselja1} investigated lattice-valued covering, or fuzzy neighboring relation arising from a given lattice-valued order and showed that every L-fuzzy covering is obtained by synthesis of crisp coverings arising from the corresponding cut orderings. Wang et al.\cite{Wang1} proposed the concepts of consistent and compatible mappings with respect to fuzzy sets and constructed a pair of lower and upper rough fuzzy approximation operators by means of the concept of fuzzy mappings. Yang et al.\cite{Yang2} presented the definition of fuzzy $\gamma$-minimal description and a novel type of fuzzy covering-based rough set model and investigate its properties. Yao et al.\cite{Yao4} introduced the concepts of fuzzy positive region reduct, lower approximation reduct and generalized fuzzy belief reduct, and investigated the relationships among these reducts. Zhang et al.\cite{Zhang6} proposed the generalized intuitionistic fuzzy rough sets based on intuitionistic fuzzy coverings and studied its properties.

\section{Preliminaries}

In this section, we briefly recall some basic concepts of
covering rough set theory, probabilistic and grade rough set theory, and fuzzy $\gamma-$covering approximation spaces.

\subsection{Covering approximation spaces}

In this section, we recall the concepts of coverings, the lower and upper approximation operators.

\begin{definition}\cite{Zakowski1}
Let $U$ be a non-empty set $($the universe of
discourse$)$. A non-empty sub-family $\mathscr{C}\subseteq
\mathscr{P} (U)$ is called a covering of $U$ if

$(1)$ every element in $\mathscr{C} $ is non-empty;

$(2)$ $\bigcup \{C \mid C \in \mathscr{C} \}=U$, where $\mathscr{P} (U)$
is the powerset of $U$.
\end{definition}

Unless stated otherwise, $U$ is a finite universe,
the covering $\mathscr{C}$ consists of finite number of sets, and the ordered pair $(U, \mathscr{C}) $ is called a covering approximation space, which is an extension of Pawlak's model using the equivalence relation. Especially, the incomplete information system is corresponding to a covering approximation space or a covering information system in fact.

We employ an example to illustrate how to construct the covering approximation space as follows.

\begin{example}
Let $U=\{x_{1},x_{2},x_{3},x_{4},x_{5},x_{6},x_{7},x_{8}\}$ be eight
cars, $C=\{price\}$ the attribute set, the domain of
$price$ is $\{high, middle, low\}$. The specialists  $A$ and $B$ are
employed to evaluate these cars and their evaluation reports are shown as follows:
\begin{eqnarray*}
high_{A}&=&\{x_{1}, x_{4}, x_{5}, x_{7}\}, middle_{A}=\{x_{2},
x_{8}\}, low_{A}=\{x_{3}, x_{6}\};\\
high_{B}&=&\{x_{1}, x_{2}, x_{4}, x_{7}, x_{8}\},
middle_{B}=\{x_{5}\}, low_{B}=\{x_{3}, x_{6}\},
\end{eqnarray*} where $high_{A}$ denotes the cars
belonging to high price by the specialist $A$, and the meanings of other
symbols are similar. Since their evaluations are of equal
importance, we should consider all their advice. Therefore, we drive the covering approximation space $(U,\mathscr{C}_{price})$, where $\mathscr{C}_{price}=\{high_{A\vee B},
middle_{A\vee B}, low_{A\vee B}\}$, and
\begin{eqnarray*}
high_{A\vee B}&=&high_{A}\cup high_{B}=\{x_{1}, x_{2}, x_{4}, x_{5},
x_{7}, x_{8}\};\\  middle_{A\vee B}&=&middle_{A}\cup
middle_{B}=\{x_{2}, x_{5}, x_{8}\};\\ low_{A\vee B}&=&low_{A}\cup
low_{B}=\{x_{3}, x_{6}\}.
\end{eqnarray*}
\end{example}

\begin{definition}\cite{Zhu1}
Let $(U,\mathscr{C})$ be a covering approximation space, where $U=\{x_{1},x_{2},...,x_{n}\}$, $\mathscr{C}=\{C_{1},C_{2},...,C_{m}\}$, and $N(x)=\bigcap\{C_{i}\mid x\in C_{i}\in \mathscr{C}\}$ for $x\in U$. Then the lower and upper approximations of $X\in P(U)$ are defined as follows:
\begin{eqnarray*}
\underline{R}(X)&=&\{x\in U|N(x)\subseteq X\};\\
\overline{R}(X)&=&\{x\in U|N(x)\cap X\neq \emptyset\}.
\end{eqnarray*}
\end{definition}

The neighborhood $N(x)$ of $x\in U$ is constructed using the covering $\mathscr{C}$, and $\{N(x)|x\in U\}$ is also a covering of $U$, but the granularity of the covering $\{N(x)|x\in U\}$ is finer than the covering $\mathscr{C}$, and the lower and upper approximation operators given by Definition 3.3 is very effective for computing the lower and upper approximations of sets in the covering approximation spaces.

\subsection{Probabilistic and grade lower and upper approximation operators}

In this section, we recall the probabilistic and grade lower and upper approximation operators in the covering approximation spaces.

\begin{definition}\cite{Yao1}
Let $(U,\mathscr{C})$ be a covering approximation space, where $U=\{x_{1},x_{2},...,x_{n}\}$, $\mathscr{C}=\{C_{1},C_{2},...,C_{m}\}$, $N(x)=\bigcap\{C_{i}\mid x\in C_{i}\in \mathscr{C}\}$ for $x\in U$, $P(X|N(x))=\frac{|X\cap N(x)|}{|N(x)|}$, and $0\leq \beta \leq \alpha \leq 1$. Then the probabilistic lower and upper approximations of the set $X\in P(U)$ are defined as follows:
\begin{eqnarray*}
 \overline{R}_{(\alpha,\beta)}(X)&=&\{x\in U\mid P(X|N(x))\geq\beta \};\\
\underline{R}_{(\alpha,\beta)}(X)&=&\{x\in U\mid P(X|N(x))\geq\alpha \}.
\end{eqnarray*}
\end{definition}

We observe that the probabilistic lower and upper approximation operators are proposed by generalizing Definition 3.3, which compute the lower and upper approximations of sets using the relative quantitative information, and the conditions of the probabilistic lower and upper approximation operators are looser than Definition 3.3.

The positive, boundary, and negative regions of the set $X\in P(U)$ using the probabilistic lower and upper approximation operators are constructed as follows:
\begin{eqnarray*}
POS_{(\alpha,\beta)}(X)&=&\{x\in U\mid P(X|N(x))\geq\alpha \};\\
BOU_{(\alpha,\beta)}(X)&=&\{x\in U\mid \beta\leq P(X|N(x))<\alpha \};\\
NEG_{(\alpha,\beta)}(X)&=&\{x\in U\mid P(X|N(x))<\beta \}.
\end{eqnarray*}

\begin{definition}\cite{Yao6}
Let $(U,\mathscr{C})$ be a covering approximation space, where $U=\{x_{1},x_{2},...,x_{n}\}$, $\mathscr{C}=\{C_{1},C_{2},...,C_{m}\}$, $N(x)$ the neighborhood of $x\in U$, and $k\in R$. Then the grade lower and upper approximations of the set $X\in P(U)$ are defined as follows:
\begin{eqnarray*}
\overline{R}_{k}(X)&=&\{x\in U\mid \Sigma_{y\in U}|X\cap N(x)|>k\};\\
\underline{R}_{k}(X)&=&\{x\in U\mid \Sigma_{y\in U}|X^{c}\cap N(x)|\leq k\}.
\end{eqnarray*}
\end{definition}

We see the grade lower and upper approximation operators are different from the probabilistic lower and upper approximation operators, which compute the lower and upper approximations of sets using the absolute quantitative information, but there are some similarities between them, and they can be transferred into each other under some conditions.

The positive, lower boundary, upper boundary, and negative regions of the set $X\in P(U)$ are computed by Definition 3.5 as follows:
\begin{eqnarray*}
POS_{k}(X)&=&\overline{R}_{k}(X)\cap \underline{R}_{k}(X);\\
NEG_{k}(X)&=&(\overline{R}_{k}(X)\cup \underline{R}_{k}(X))^{c};\\
LBO_{k}(X)&=&\underline{R}_{k}(X)-\overline{R}_{k}(X);\\
UBO_{k}(X)&=&\overline{R}_{k}(X)-\underline{R}_{k}(X);\\
BOU_{k}(X)&=&LBO_{R_{k}}(X)\cup UBO_{R_{k}}(X).
\end{eqnarray*}

\subsection{Fuzzy $\gamma-$covering approximation spaces}

In this section, we recall some concepts of fuzzy covering approximation spaces.

\begin{definition}\cite{Zadeh1}
Let $\mu_{A}$ be a mapping from $U$ to $[0,1]$ such as $\mu_{A}:
U\longrightarrow [0,1]:$ $ x\longrightarrow \mu_{A}(x),$ where $x\in U$, and $\mu_{A}$ is the membership function. Then $A$ is referred to as a fuzzy set.
\end{definition}

We denote the family of all fuzzy subsets of $U$ and $\mu_{A}(x)$ as $\mathscr{F}(U)$ and $A(x)$, respectively, for simplicity. For any $A,B\in \mathscr{F}(U)$, if $A(x)\leq B(x)$ for any $x\in U$, then we say $A$ is contained in $B$, denoted as $A\subseteq B$. Especially, $A=B$ if and only if $A\subseteq B$ and $B\subseteq A$. We also have $(A\cup B)(x)=A(x)\vee B(x)$, $(A\cap B)(x)=A(x)\wedge B(x)$, and $A^{c}(x)=1-A(x)$ for any $x\in U$.

We also employ an example to illustrate the union, intersection, and complement of fuzzy sets as follows.

\begin{example}(Continuation from Example 3.2)
Let $A$ and $B$ be fuzzy subsets of the universe $U$ as follows:
\begin{eqnarray*}
A&=&\frac{1}{x_{1}}+\frac{0.6}{x_{2}}+\frac{0}{x_{3}}
+\frac{0.8}{x_{4}}+\frac{1}{x_{5}}
+\frac{0}{x_{6}}+\frac{0.8}{x_{7}}+\frac{1}{x_{8}}
;\\
B&=&\frac{1}{x_{1}}+\frac{0}{x_{2}}+\frac{0.6}{x_{3}}
+\frac{1}{x_{4}}+\frac{0}{x_{5}}
+\frac{0.8}{x_{6}}+\frac{1}{x_{7}}+\frac{0.8}{x_{8}}.
\end{eqnarray*}

By Definition 3.6, we have that $A(x_{1})=1,A(x_{2})=0.6,A(x_{3})=0,A(x_{4})=0.8,
A(x_{5})=1,A(x_{6})=0,A(x_{7})=0.8,A(x_{8})=1,$ $B(x_{1})=1,B(x_{2})=0,B(x_{3})=0.6,B(x_{4})=1,
B(x_{5})=0,B(x_{6})=0.8,B(x_{7})=1,$ and $B(x_{8})=0.8.$ We also have $A\cap B$, $A\cup B$, and $A^{c}$ as follows:
\begin{eqnarray*}
A\cap B&=&\frac{1}{x_{1}}+\frac{0}{x_{2}}+\frac{0}{x_{3}}
+\frac{0.8}{x_{4}}+\frac{0}{x_{5}}
+\frac{0}{x_{6}}+\frac{0.8}{x_{7}}+\frac{0.8}{x_{8}}
;\\
A\cup B&=&\frac{1}{x_{1}}+\frac{0.6}{x_{2}}+\frac{0.6}{x_{3}}
+\frac{1}{x_{4}}+\frac{1}{x_{5}}
+\frac{0.8}{x_{6}}+\frac{1}{x_{7}}+\frac{1}{x_{8}};\\
A^{c}&=&\frac{0}{x_{1}}+\frac{0.4}{x_{2}}+\frac{1}{x_{3}}
+\frac{0.2}{x_{4}}+\frac{0}{x_{5}}
+\frac{1}{x_{6}}+\frac{0.2}{x_{7}}+\frac{0}{x_{8}}.
\end{eqnarray*}
\end{example}

\begin{definition}\cite{Ma1}
A fuzzy $\gamma-$covering of $U$ is a collection of fuzzy sets
$\mathscr{C}^{\ast}\subseteq \mathscr{F}(U)$  which satisfies

$(1)$ every fuzzy set $C^{\ast}\in \mathscr{C}^{\ast}$ is non-empty,
i.e., $C^{\ast}\neq\emptyset$;

$(2)$ $ \forall x\in U, \bigvee_{C^{\ast}\in
\mathscr{C}^{\ast}}C^\ast(x)\geq \gamma$.
\end{definition}

Unless stated otherwise, $U$ is a finite universe, the fuzzy covering
$\mathscr{C}^{\ast}$ consists of finite number of sets, and the ordered pair $(U, \mathscr{C}^{\ast}) $ is called a $\gamma-$fuzzy covering approximation space, as an extension of the covering approximation space.

\begin{example}
(Continuation from Example 3.2) To evaluate these cars, specialists $A$ and $B$ are
employed and their evaluation reports are shown as follows:
\begin{eqnarray*}
high^{\ast}_{A}&=&\frac{1}{x_{1}}+\frac{0.7}{x_{2}}+\frac{0}{x_{3}}
+\frac{0.9}{x_{4}}+\frac{0.9}{x_{5}}
+\frac{0}{x_{6}}+\frac{0.9}{x_{7}}+\frac{0.6}{x_{8}};\\
middle^{\ast}_{A}&=&\frac{0.6}{x_{1}}+\frac{0.9}{x_{2}}+\frac{0.4}{x_{3}}
+\frac{0.4}{x_{4}}+\frac{0.5}{x_{5}}
+\frac{0.5}{x_{6}}+\frac{0.5}{x_{7}}+\frac{0.9}{x_{8}};\\
low^{\ast}_{A}&=&\frac{0}{x_{1}}+\frac{0.5}{x_{2}}+\frac{0.9}{x_{3}}
+\frac{0}{x_{4}}+\frac{0.5}{x_{5}}
+\frac{0.9}{x_{6}}+\frac{0}{x_{7}}+\frac{0.5}{x_{8}};\\
high^{\ast}_{B}&=&\frac{0.9}{x_{1}}+\frac{0.7}{x_{2}}+\frac{0}{x_{3}}
+\frac{0.9}{x_{4}}+\frac{0.9}{x_{5}}
+\frac{0}{x_{6}}+\frac{0.9}{x_{7}}+\frac{0.8}{x_{8}};\\
middle^{\ast}_{B}&=&\frac{0.6}{x_{1}}+\frac{0.9}{x_{2}}+\frac{0.4}{x_{3}}
+\frac{0.4}{x_{4}}+\frac{0.5}{x_{5}}
+\frac{0.7}{x_{6}}+\frac{0.5}{x_{7}}+\frac{1}{x_{8}};\\
low^{\ast}_{B}&=&\frac{0}{x_{1}}+\frac{0.5}{x_{2}}+\frac{0.9}{x_{3}}
+\frac{0}{x_{4}}+\frac{0.5}{x_{5}}
+\frac{0.9}{x_{6}}+\frac{0}{x_{7}}+\frac{0.5}{x_{8}},
\end{eqnarray*}
where $high^{\ast}_{A}$ is the membership degree of each car belonging to
the high price by the specialist $A$. The meanings of the other
symbols are similar. Then we obtain a $0.9-$fuzzy
covering approximation space $(U,\mathscr{C}^{\ast}_{price})$, where
$\mathscr{C}^{\ast}_{price}=\{C^{\ast}_{high}, C^{\ast}_{middle}, C^{\ast}_{low}\}$, and
\begin{eqnarray*}
C^{\ast}_{high}&=&high^{\ast}_{A}\cup
high^{\ast}_{B}=\frac{1}{x_{1}}+\frac{0.7}{x_{2}}+\frac{0}{x_{3}}
+\frac{0.9}{x_{4}}+\frac{0.9}{x_{5}}
+\frac{0}{x_{6}}+\frac{0.9}{x_{7}}+\frac{0.8}{x_{8}};\\
C^{\ast}_{middle}&=&middle^{\ast}_{A}\cup
middle^{\ast}_{B}=\frac{0.6}{x_{1}}+\frac{0.9}{x_{2}}+\frac{0.4}{x_{3}}
+\frac{0.4}{x_{4}}+\frac{0.5}{x_{5}}
+\frac{0.7}{x_{6}}+\frac{0.5}{x_{7}}+\frac{1}{x_{8}};\\
C^{\ast}_{low}&=&low^{\ast}_{A}\cup
low^{\ast}_{B}=\frac{0}{x_{1}}+\frac{0.5}{x_{2}}+\frac{0.9}{x_{3}}
+\frac{0}{x_{4}}+\frac{0.5}{x_{5}}
+\frac{0.9}{x_{6}}+\frac{0}{x_{7}}+\frac{0.5}{x_{8}}.
\end{eqnarray*}
\end{example}

It is obvious that we can construct a fuzzy $\gamma-$covering of the universe
with an attribute. Since the fuzzy covering rough set theory is
very effective to handle uncertain information, the investigation of this
theory becomes an important task in rough set theory.

\section{Double-quantitative approximation operators}

In this section, we provide the concepts of the fuzzy $\gamma-$covering based probabilistic approximation operators, grade approximation operators, and double-quantitative approximation operators.

\subsection{Probabilistic lower and upper approximation operators}

In this section, we recall the concept of the fuzzy $\gamma-$neighborhood $\widetilde{N}_{x}^{\gamma}$ of $x\in U$ as follows.

\begin{definition}\cite{Ma1}
Let $(U,\mathscr{C}^{\ast})$ be a fuzzy $\gamma-$covering approximation space, where $U=\{x_{1},x_{2},...,x_{n}\}$, and $\mathscr{C}^{\ast}=\{C^{\ast}_{1},C^{\ast}_{2},...,C^{\ast}_{m}\}$, and $\gamma\in (0,1]$. Then the fuzzy $\gamma-$neighborhood $\widetilde{N}_{x}^{\gamma}$ of $x\in U$ is defined as follows:
$$\widetilde{N}_{x}^{\gamma}=\bigcap\{C^{\ast}_{i}\in \mathscr{C}^{\ast}\mid C^{\ast}_{i}(x)\geq \gamma\}.$$
\end{definition}

The concept of the fuzzy $\gamma-$neighborhood operator $\widetilde{N}_{x}^{\gamma}$ is an extension of the classical neighborhood $N(x)$ in the fuzzy $\gamma-$covering approximation space, which will be applied to compute the fuzzy $\gamma-$covering based probabilistic lower and upper approximations of fuzzy sets. In what follows, we denote $\mathscr{C}^{\ast}$ and $C^{\ast}_{i}$ as $\mathscr{C}$ and $C_{i}$, respectively, for simplicity.

\begin{definition}
Let $(U,\mathscr{C})$ be a fuzzy $\gamma-$covering approximation space, where $U=\{x_{1},x_{2},...,x_{n}\}$,
$\mathscr{C}=\{C_{1},C_{2},...,C_{m}\}$, and $\gamma\in (0,1]$. Then the conditional probability $P(X|\widetilde{N}_{x}^{\gamma})$ of the fuzzy event $X\in \mathscr{F}(U)$
given the description $\widetilde{N}_{x}^{\gamma}$ is defined as follows:
\begin{eqnarray*}
 P(X|\widetilde{N}_{x}^{\gamma})=\frac{\Sigma_{y\in U}(X\cap\widetilde{N}_{x}^{\gamma})(y)}{\Sigma_{y\in U}\widetilde{N}_{x}^{\gamma}(y)}.
\end{eqnarray*}
\end{definition}

The concept of the conditional probability $P(X|\widetilde{N}_{x}^{\gamma})$ of the fuzzy event $X\in \mathscr{F}(U)$ is an generalization of the conditional probability $P(X|N(x))$ of the event $X\in P(U)$, which is helpful for studying the fuzzy $\gamma-$covering approximation space.

In what follows, we propose the concept of the fuzzy $\gamma-$covering based probabilistic lower and upper approximation operators in the fuzzy $\gamma-$covering approximation space.

\begin{definition}
Let $(U,\mathscr{C})$ be a fuzzy $\gamma-$covering approximation space, where $U=\{x_{1},x_{2},...,x_{n}\}$, $\mathscr{C}=\{C_{1},C_{2},...,C_{m}\}$, and $0\leq \beta \leq \alpha\leq 1$. Then the fuzzy $\gamma-$covering based probabilistic lower and upper approximations of the fuzzy set $X\in \mathscr{F}(U)$ are defined as follows:
\begin{eqnarray*}
\overline{FR}_{(\alpha,\beta)}(X)&=&\{x\in U\mid P(X|\widetilde{N}_{x}^{\gamma})\geq\beta \};\\
\underline{FR}_{(\alpha,\beta)}(X)&=&\{x\in U\mid P(X|\widetilde{N}_{x}^{\gamma})\geq\alpha \}.
\end{eqnarray*}
\end{definition}

The fuzzy $\gamma-$covering based probabilistic lower and upper approximation operators $\overline{FR}_{(\alpha,\beta)}(X)$ and $\underline{FR}_{(\alpha,\beta)}(X)$ for the fuzzy set $X\in \mathscr{F}(U)$ given by Definition 4.3 are extensions of the probabilistic lower and upper approximation operators $\overline{R}_{(\alpha,\beta)}(X)$ and $\underline{R}_{(\alpha,\beta)}(X)$ for the set $X\in P(U)$ given by Definition 3.4, which construct the lower and upper approximations of fuzzy sets using the relative quantitative information.

\begin{example}
(Continuation from Example 3.9) Taking $X=\frac{0.6}{x_{1}}+\frac{0.5}{x_{2}}+\frac{0.7}{x_{3}}
+\frac{0.8}{x_{4}}+\frac{0.5}{x_{5}}
+\frac{0.6}{x_{6}}+\frac{0}{x_{7}}+\frac{0.2}{x_{8}}$, $\alpha=0.75,$ and $\beta=0.25$, we have the fuzzy $\gamma-$covering based probabilistic lower and upper approximations of $X$ as follows:
\begin{eqnarray*}
\underline{FR}_{(\alpha,\beta)}(X)=\{x_{3},x_{6}\}\text{ and }
\overline{FR}_{(\alpha,\beta)}(X)=\{x_{1},x_{2},x_{3},x_{4},x_{5},x_{6},x_{7},x_{8}\}.
\end{eqnarray*}
\end{example}

\begin{definition}
Let $(U,\mathscr{C})$ be a fuzzy $\gamma-$covering approximation space, where $U=\{x_{1},x_{2},...,x_{n}\}$, $\mathscr{C}=\{C_{1},C_{2},...,C_{m}\}$, and $0\leq \beta \leq \alpha \leq 1$. Then the fuzzy $\gamma-$covering based probabilistic positive, boundary, and negative regions of the fuzzy set $X\in \mathscr{F}(U)$ are defined as follows:
\begin{eqnarray*}
\widetilde{POS}_{(\alpha,\beta)}(X)&=&\{x\in U\mid P(X|\widetilde{N}_{x}^{\gamma})\geq\alpha \};\\
\widetilde{BOU}_{(\alpha,\beta)}(X)&=&\{x\in U\mid \beta\leq P(X|\widetilde{N}_{x}^{\gamma})<\alpha \};\\
\widetilde{NEG}_{(\alpha,\beta)}(X)&=&\{x\in U\mid P(X|\widetilde{N}_{x}^{\gamma})<\beta \}.
\end{eqnarray*}
\end{definition}

The the fuzzy $\gamma-$covering based probabilistic positive, boundary, and negative regions $\widetilde{POS}_{(\alpha,\beta)}(X),$ $\widetilde{BOU}_{(\alpha,\beta)}(X), \widetilde{NEG}_{(\alpha,\beta)}(X)$ of the fuzzy set $X\in \mathscr{F}(U)$ in the fuzzy $\gamma-$covering approximation space are generalizations of the probabilistic positive, boundary, and negative regions $POS_{(\alpha,\beta)}(X), BOU_{(\alpha,\beta)}(X)$ and $NEG_{(\alpha,\beta)}(X)$ of the set $X\in P(U)$ in the covering approximation spaces.

\begin{example}(Continuation from Example 4.4) Taking $\alpha=0.75,$ and $\beta=0.25$, we have the fuzzy $\gamma-$covering based probabilistic positive, boundary, and negative regions of $X$ as follows:
\begin{eqnarray*}
\widetilde{POS}_{(\alpha,\beta)}(X)&=&\{x\in U\mid P(X|\widetilde{N}_{x}^{\gamma})\geq\alpha \}=\{x_{3},x_{6}\};\\
\widetilde{BOU}_{(\alpha,\beta)}(X)&=&\{x\in U\mid \beta\leq P(X|\widetilde{N}_{x}^{\gamma})<\alpha \}=\{x_{1},x_{2},x_{4},x_{5},x_{7},x_{8}\};\\
\widetilde{NEG}_{(\alpha,\beta)}(X)&=&\{x\in U\mid P(X|\widetilde{N}_{x}^{\gamma})<\beta \}=\emptyset.
\end{eqnarray*}
\end{example}

We study the basic properties of the fuzzy $\gamma-$covering based lower and upper approximations of sets as follows.

\begin{theorem}
Let $(U,\mathscr{C})$ be a fuzzy $\gamma-$covering approximation space, where $U=\{x_{1},x_{2},...,x_{n}\}$, $\mathscr{C}=\{C_{1},C_{2},...,C_{m}\}$, $0\leq \beta< \alpha\leq 1$, and $X,Y\in \mathscr{F}(U)$. Then\\
$(1) \underline{FR}_{(\alpha,\beta)}(U)=U;\overline{FR}_{(\alpha,\beta)}(\emptyset)= \emptyset;\\
(2) X\subseteq Y\Rightarrow\overline{FR}_{(\alpha,\beta)}(X)\subseteq \overline{FR}_{(\alpha,\beta)}(Y);\\
(3) X\subseteq Y\Rightarrow\underline{FR}_{(\alpha,\beta)}(X)\subseteq \underline{FR}_{(\alpha,\beta)}(Y);\\
(4) \overline{FR}_{(\alpha,\beta)}(X)\cup \overline{FR}_{(\alpha,\beta)}(Y)\subseteq \overline{FR}_{(\alpha,\beta)}(X\cup Y);\\
(5)\underline{FR}_{(\alpha,\beta)}(X)\cup \underline{FR}_{(\alpha,\beta)}(Y)\subseteq \underline{FR}_{(\alpha,\beta)}(X\cup Y);\\
(6) \overline{FR}_{(\alpha,\beta)}(X\cap Y)\subseteq \overline{FR}_{(\alpha,\beta)}(X)\cap \overline{FR}_{(\alpha,\beta)}(Y);\\
(7)\underline{FR}_{(\alpha,\beta)}(X\cap Y)\subseteq \underline{FR}_{(\alpha,\beta)}(X)\cap \underline{FR}_{(\alpha,\beta)}(Y);\\
(8) \alpha_{1} \leq \alpha_{2},\beta_{1} \leq \beta_{2} \Rightarrow\underline{FR}_{(\alpha_{2},\beta_{2})}(X)\subseteq \underline{FR}_{(\alpha_{1},\beta_{1})}(X);\\
(9) \alpha_{1} \leq \alpha_{2},\beta_{1} \leq \beta_{2} \Rightarrow\overline{FR}_{(\alpha_{2},\beta_{2})}(X)\subseteq \overline{FR}_{(\alpha_{1},\beta_{1})}(X).$
\end{theorem}

\noindent\textbf{Proof.}
(1) For any $x\in U$, we have $P(U|\widetilde{N}_{x}^{\gamma})=\frac{\Sigma_{y\in U}(U\cap\widetilde{N}_{x}^{\gamma})(y)}{\Sigma_{y\in U}\widetilde{N}_{x}^{\gamma}(y)}=1\geq \alpha$ and $P(\emptyset|\widetilde{N}_{x}^{\gamma})=\frac{\Sigma_{y\in U}(\emptyset\cap\widetilde{N}_{x}^{\gamma})(y)}{\Sigma_{y\in U}\widetilde{N}_{x}^{\gamma}(y)}=0< \beta$ by Definition 4.2. So $\underline{FR}_{(\alpha,\beta)}(U)=U$ and $\overline{FR}_{(\alpha,\beta)}(\emptyset)=\emptyset$.

(2) For $x_{0}\in \overline{R}_{(\alpha,\beta)}(X)$, we have $P(X|\widetilde{N}_{x_{0}}^{\gamma})=\frac{\Sigma_{y\in U}(X\cap\widetilde{N}_{x_{0}}^{\gamma})(y)}{\Sigma_{y\in U}\widetilde{N}_{x_{0}}^{\gamma}(y)}\geq\beta$. Since $X\subseteq Y$, we have $\Sigma_{y\in U}(X\cap\widetilde{N}_{x_{0}}^{\gamma})(y)\leq \Sigma_{y\in U}(Y\cap\widetilde{N}_{x_{0}}^{\gamma})(y)$. It follows $P(X|\widetilde{N}_{x_{0}}^{\gamma})\leq P(Y|\widetilde{N}_{x_{0}}^{\gamma})$. So $x_{0}\in \overline{R}_{(\alpha,\beta)}(Y)$. Therefore, $\overline{FR}_{(\alpha,\beta)}(X)\subseteq \overline{FR}_{(\alpha,\beta)}(Y)$.

(3) For $x_{0}\in \underline{FR}_{(\alpha,\beta)}(X)$, we have $P(X|\widetilde{N}_{x_{0}}^{\gamma})=\frac{\Sigma_{y\in U}(X\cap\widetilde{N}_{x_{0}}^{\gamma})(y)}{\Sigma_{y\in U}\widetilde{N}_{x_{0}}^{\gamma}(y)}\geq\alpha$. Since $X\subseteq Y$, we have $\Sigma_{y\in U}(X\cap\widetilde{N}_{x_{0}}^{\gamma})(y)\leq \Sigma_{y\in U}(Y\cap\widetilde{N}_{x_{0}}^{\gamma})(y)$. It follows $P(X|\widetilde{N}_{x_{0}}^{\gamma})\leq P(Y|\widetilde{N}_{x_{0}}^{\gamma})$. We obtain $x_{0}\in \underline{FR}_{(\alpha,\beta)}(Y)$. So $\underline{FR}_{(\alpha,\beta)}(X)\subseteq \underline{FR}_{(\alpha,\beta)}(Y)$.

(4) By Theorem 4.7(2), we have $\overline{FR}_{(\alpha,\beta)}(X)\subseteq \overline{FR}_{(\alpha,\beta)}(X\cup Y)$ and $\overline{FR}_{(\alpha,\beta)}(Y)\subseteq \overline{FR}_{(\alpha,\beta)}(X\cup Y)$ for $X,Y\in \mathscr{F}(U)$. Therefore, $\overline{FR}_{(\alpha,\beta)}(X)\cup\overline{FR}_{(\alpha,\beta)}(Y)\subseteq \overline{FR}_{(\alpha,\beta)}(X\cup Y)$

(5) By Theorem 4.7(2), we have $\underline{FR}_{(\alpha,\beta)}(X)\subseteq \underline{FR}_{(\alpha,\beta)}(X\cup Y)$ and $\underline{FR}_{(\alpha,\beta)}(Y)\subseteq \underline{FR}_{(\alpha,\beta)}(X\cup Y)$ for $X,Y\in \mathscr{F}(U)$. So $\underline{FR}_{(\alpha,\beta)}(X)\cup\underline{FR}_{(\alpha,\beta)}(Y)\subseteq \underline{FR}_{(\alpha,\beta)}(X\cup Y)$.

(6) and (7) The proof is similar to Theorem 4.7(3) and (4).

(8) By Definition 4.3, we have $\underline{FR}_{(\alpha_{1},\beta_{1})}(X)=\{x\in U\mid P(X|\widetilde{N}_{x}^{\gamma})\geq\alpha_{1} \}$ and $\underline{FR}_{(\alpha_{2},\beta_{2})}(X)=\{x\in U\mid P(X|\widetilde{N}_{x}^{\gamma})\geq\alpha_{2} \}$. If $P(X|\widetilde{N}_{z}^{\gamma})\geq\alpha_{2}$ for $z\in U$, we have $P(X|\widetilde{N}_{z}^{\gamma})\geq\alpha_{1}$ for $z\in U$ since $\alpha_{1} \leq \alpha_{2}$. Therefore, $\underline{FR}_{(\alpha_{2},\beta_{2})}(X)\subseteq \underline{FR}_{(\alpha_{1},\beta_{1})}(X)$.

(9) By Definition 4.3, we have $\overline{FR}_{(\alpha_{1},\beta_{1})}(X)=\{x\in U\mid P(X|\widetilde{N}_{x}^{\gamma})\geq\beta_{1} \}$ and $\overline{FR}_{(\alpha_{2},\beta_{2})}(X)=\{x\in U\mid P(X|\widetilde{N}_{x}^{\gamma})\geq\beta_{2} \}$. If $P(X|\widetilde{N}_{z}^{\gamma})\geq\beta_{2}$ for $z\in U$, we have $P(X|\widetilde{N}_{z}^{\gamma})\geq\beta_{1}$ for $z\in U$ since $\beta_{1} \leq \beta_{2}$. So $\overline{FR}_{(\alpha_{2},\beta_{2})}(X)\subseteq \overline{FR}_{(\alpha_{1},\beta_{1})}(X)$.
$\Box$

\subsection{Grade lower and upper approximation operators}

In this section, we propose the fuzzy $\gamma-$covering based grade lower and upper approximation operators for the fuzzy $\gamma-$covering approximation space.

\begin{definition}
Let $(U,\mathscr{C})$ be a fuzzy $\gamma-$covering approximation space, where $U=\{x_{1},x_{2},...,x_{n}\}$, $\mathscr{C}=\{C_{1},C_{2},...,C_{m}\}$, $\gamma\in (0,1]$, and $k\in R$. Then the fuzzy $\gamma-$covering based grade lower and upper approximations of the fuzzy set $X\in \mathscr{F}(U)$ are defined as follows:
\begin{eqnarray*}
\overline{GR}_{k}(X)&=&\{x\in U\mid \Sigma_{y\in U}(X\cap \widetilde{N}_{x}^{\gamma})(y)>k\};\\
\underline{GR}_{k}(X)&=&\{x\in U\mid \Sigma_{y\in U}[\widetilde{N}_{x}^{\gamma}(y)-(X\cap \widetilde{N}_{x}^{\gamma})(y)]\leq k\}.
\end{eqnarray*}
\end{definition}

The fuzzy $\gamma-$covering based grade lower and upper approximation operators $\overline{GR}_{k}(X)$ and $\underline{GR}_{k}(X)$ for the fuzzy set $X\in \mathscr{F}(U)$ given by Definition 4.8 are extensions of the grade lower and upper approximation operators $\overline{R}_{k}(X)$ and $\underline{R}_{k}(X)$ for the set $X\in P(U)$ given by Definition 3.5, which construct the lower and upper approximations of fuzzy sets using the absolute quantitative information.

We employ an example to illustrate the construction of the fuzzy $\gamma-$covering based grade lower and upper approximations of sets as follows.

\begin{example}
(Continuation from Example 4.4) Taking $k=2$, we have the fuzzy $\gamma-$covering based grade lower and upper approximations of the fuzzy set $X$ as follows:
\begin{eqnarray*}
\overline{GR}_{2}(X)=\{x_{1},x_{2},x_{3},x_{4},x_{5},x_{6},x_{7},x_{8}\} \text{ and }
\underline{GR}_{2}(X)=\{x_{2},x_{3},x_{6},x_{8}\}.
\end{eqnarray*}
\end{example}

\begin{definition}
Let $(U,\mathscr{C})$ be a fuzzy $\gamma-$covering approximation space, where $U=\{x_{1},x_{2},...,x_{n}\}$, $\mathscr{C}=\{C_{1},C_{2},...,C_{m}\}$, $\gamma\in (0,1]$, and $k\in R$. Then the fuzzy $\gamma-$covering based grade positive, boundary, and negative regions of the fuzzy set $X\in \mathscr{F}(U)$ are defined as follows:
\begin{eqnarray*}
\widetilde{POS}_{k}(X)&=&\overline{GR}_{k}(X)\cap \underline{GR}_{k}(X);\\
\widetilde{NEG}_{k}(X)&=&(\overline{GR}_{k}(X)\cup \underline{GR}_{k}(X))^{c};\\
\widetilde{LBO}_{k}(X)&=&\underline{GR}_{k}(X)-\overline{GR}_{k}(X);\\
\widetilde{UBO}_{k}(X)&=&\overline{GR}_{k}(X)-\underline{GR}_{k}(X);\\
\widetilde{BOU}_{k}(X)&=&\widetilde{LBO}_{k}(X)\cup \widetilde{UBO}_{k}(X).
\end{eqnarray*}
\end{definition}

The fuzzy $\gamma-$covering based grade positive, lower and upper boundary, and negative regions $\widetilde{POS}_{k}(X),$ $\widetilde{NEG}_{k}(X),\widetilde{LBO}_{k}(X),$ $\widetilde{UBO}_{k}(X)$, and $\widetilde{BOU}_{k}(X)$ of the fuzzy set $X\in \mathscr{F}(U)$ in the fuzzy $\gamma-$covering approximation space are generalizations of the grade positive, lower and upper boundary, and negative regions $POS_{k}(X), NEG_{k}(X),LBO_{k}(X),$ $UBO_{k}(X),$ and $BOU_{k}(X)$ of the set $X\in P(U)$ in the covering approximation spaces.

\begin{example}
(Continuation from Example 4.4)
Taking $k=2$, we have the fuzzy $\gamma-$covering based grade positive, lower and upper boundary, and negative regions of the fuzzy set $X$ as follows:
\begin{eqnarray*}
\widetilde{POS}_{2}(X)&=&\overline{GR}_{2}(X)\cap \underline{GR}_{2}(X)=\{x_{2},x_{3},x_{6},x_{8}\};\\
\widetilde{NEG}_{2}(X)&=&(\overline{GR}_{2}(X)\cup \underline{GR}_{2}(X))^{c}=\emptyset;\\
\widetilde{LBO}_{2}(X)&=&\underline{GR}_{2}(X)-\overline{GR}_{2}(X)=\emptyset;\\
\widetilde{UBO}_{2}(X)&=&\overline{GR}_{2}(X)-\underline{GR}_{2}(X)=\{x_{1},x_{5},x_{6},x_{7}\};\\
\widetilde{BOU}_{2}(X)&=&\widetilde{LBO}_{2}(X)\cup \widetilde{UBO}_{2}(X)=\{x_{1},x_{5},x_{6},x_{7}\}.
\end{eqnarray*}
\end{example}

We present the basic properties of the fuzzy $\gamma-$covering based grade lower and upper approximation operators as follows.

\begin{theorem}
Let $(U,\mathscr{C})$ be a fuzzy $\gamma-$covering approximation space, where $U=\{x_{1},x_{2},...,x_{n}\}$, $\mathscr{C}=\{C_{1},C_{2},...,C_{m}\}$, $k,k_{1},k_{2}\in R$, and $X,Y\in \mathscr{F}(U)$. Then\\
$(1) \underline{GR}_{k}(U)=U;\overline{GR}_{k}(\emptyset)= \emptyset;\\
(2) X\subseteq Y\Rightarrow\overline{GR}_{k}(X)\subseteq \overline{GR}_{k}(Y);\\
(3) X\subseteq Y\Rightarrow\underline{GR}_{k}(X)\subseteq \underline{GR}_{k}(Y);\\
(4) \overline{GR}_{k}(X)\cup \overline{GR}_{k}(Y)\subseteq \overline{GR}_{k}(X\cup Y);\\
(5) \underline{GR}_{k}(X)\cup \underline{GR}_{k}(Y)\subseteq \underline{GR}_{k}(X\cup Y);\\
(6) \overline{GR}_{k}(X\cap Y)\subseteq \overline{GR}_{k}(X)\cap \overline{GR}_{k}(Y);\\
(7) \underline{GR}_{k}(X\cap Y)\subseteq \underline{GR}_{k}(X)\cap \underline{GR}_{k}(Y);\\
(8) k_{1} \leq k_{2} \Rightarrow\overline{GR}_{k_{1}}(X)\subseteq \overline{GR}_{k_{2}}(X);\\
(9) k_{1} \leq k_{2}\Rightarrow\underline{GR}_{k_{2}}(X)\subseteq \underline{GR}_{k_{1}}(X).$
\end{theorem}

\noindent\textbf{Proof.}
(1) By Definition 4.8, $\Sigma_{y\in U}(U^{c}\cap \widetilde{N}_{x}^{\gamma})(y)=0\leq k$ and $\Sigma_{y\in U}(\emptyset\cap \widetilde{N}_{x}^{\gamma})(y)=0\leq k$ for any $x\in U$. It follows that $\underline{GR}_{k}(U)=U$ and $\overline{GR}_{k}(\emptyset)=\emptyset$.

(2) For any $x_{0}\in \overline{GR}_{k}(X)$, we have $\Sigma_{y\in U}(X\cap \widetilde{N}_{x_{0}}^{\gamma})(y)\leq \Sigma_{y\in U}(Y\cap \widetilde{N}_{x_{0}}^{\gamma})(y)$.
Since $\Sigma_{y\in U}(X\cap \widetilde{N}_{x_{0}}^{\gamma})(y)>k$. It follows $\Sigma_{y\in U}(Y\cap \widetilde{N}_{x_{0}}^{\gamma})(y)>k$. So $x_{0}\in \overline{R}_{k}(Y)$.

(3) For any $x_{0}\in \underline{GR}_{k}(X)$, we have $\Sigma_{y\in U}(X^{c}\cap \widetilde{N}_{x_{0}}^{\gamma})(y)> \Sigma_{y\in U}(Y^{c}\cap \widetilde{N}_{x_{0}}^{\gamma})(y)$.
Since $\Sigma_{y\in U}(X\cap \widetilde{N}_{x_{0}}^{\gamma})(y)<k$. It follows $\Sigma_{y\in U}(Y^{c}\cap \widetilde{N}_{x_{0}}^{\gamma})(y)<k$. Therefore, $x_{0}\in \underline{GR}_{k}(Y)$.

(4) By Theorem 4.12(2), we have $\underline{GR}_{k}(X)\subseteq \underline{GR}_{k}(X\cup Y)$
and $\underline{GR}_{k}(Y)\subseteq \underline{GR}_{k}(X\cup Y)$ for $X,Y\in \mathscr{F}(U)$. It follows that $\underline{GR}_{k}(X)\cup \underline{GR}_{k}(Y)\subseteq \underline{GR}_{k}(X\cup Y)$.

(5) By Theorem 4.12(3), we have $\overline{GR}_{k}(X)\subseteq \overline{GR}_{k}(X\cup Y)$
and $\overline{GR}_{k}(Y)\subseteq \overline{GR}_{k}(X\cup Y)$ for $X,Y\in \mathscr{F}(U)$. It follows that $\overline{GR}_{k}(X)\cup \overline{GR}_{k}(Y)\subseteq \overline{GR}_{k}(X\cup Y)$.

(6),(7) The proof is similar to Theorem 4.12(4) and (5).

(8) By Definition 4.8, we have $\overline{GR}_{k_{1}}(X)=\{x\in U\mid \Sigma_{y\in U}(X^{c}\cap \widetilde{N}_{x}^{\gamma})(y)>k_{1}\}$ and $\overline{GR}_{k_{2}}(X)=\{x\in U\mid \Sigma_{y\in U}(X^{c}\cap \widetilde{N}_{x}^{\gamma})(y)>k_{2}\}$ for $X\in \mathscr{F}(U)$. For any $z\in \overline{GR}_{k_{1}}(X)$, we have $\Sigma_{y\in U}(X^{c}\cap \widetilde{N}_{x}^{\gamma})(z)>k_{1}\geq k_{2}$. It follows that $z\in \overline{GR}_{k_{2}}(X)$. Therefore, $\overline{GR}_{k_{1}}(X)\subseteq \overline{GR}_{k_{2}}(X)$.

(9) By Theorem 4.12, we have $\underline{GR}_{k_{1}}(X)=\{x\in U\mid \Sigma_{y\in U}(X\cap \widetilde{N}_{x}^{\gamma})(y)<k_{1}\}$ and $\underline{GR}_{k_{2}}(X)=\{x\in U\mid \Sigma_{y\in U}(X^{c}\cap \widetilde{N}_{x}^{\gamma})(y)<k_{2}\}$ for $X\in \mathscr{F}(U)$. For any $z\in \underline{GR}_{k_{1}}(X)$, we have $\Sigma_{y\in U}(X^{c}\cap \widetilde{N}_{x}^{\gamma})(z)<k_{1}\leq k_{2}$. It follows that $z\in \underline{GR}_{k_{2}}(X)$. Therefore, $\underline{GR}_{k_{2}}(X)\subseteq \underline{GR}_{k_{1}}(X)$.
$\Box$

In what follows, we discuss the relationship between the fuzzy $\gamma-$covering based probabilistic lower and upper approximation operators and the fuzzy $\gamma-$covering based grade lower and upper approximation operators.

\begin{theorem}
Let $(U,\mathscr{C})$ be a fuzzy $\gamma-$covering approximation space, where $U=\{x_{1},x_{2},...,x_{n}\}$, $\mathscr{C}=\{C_{1},C_{2},...,C_{m}\}$, $X\in \mathscr{F}(U)$, $\gamma\in (0,1]$, and $k\in R$. Then
\begin{eqnarray*}
 \overline{FR}_{(\alpha,\beta)}(X)&=&\{x\in U\mid  \Sigma_{y\in U}(X\cap \widetilde{N}_{x}^{\gamma})(y)\geq\beta\Sigma_{y\in U}\widetilde{N}_{x}^{\gamma}(y)\};\\
 \underline{FR}_{(\alpha,\beta)}(X)&=&\{x\in U\mid \Sigma_{y\in U}[\widetilde{N}_{x}^{\gamma})(y)-(X\cap \widetilde{N}_{x}^{\gamma})(y)]\leq \Sigma_{y\in U}\widetilde{N}_{x}^{\gamma}(y)\}-\alpha\Sigma_{y\in U}\widetilde{N}_{x}^{\gamma}(y)\}\}.
\end{eqnarray*}
\end{theorem}

\noindent\textbf{Proof.} By Definition 4.3, we have $\overline{FR}_{(\alpha,\beta)}(X)=\{x\in U\mid P(X|\widetilde{N}_{x}^{\gamma})\geq\beta \}$, which implies that $P(X|\widetilde{N}_{x}^{\gamma})\geq\beta$ for $x\in \overline{FR}_{(\alpha,\beta)}(X)$. It follows that $P(X|\widetilde{N}_{x}^{\gamma})=\frac{\Sigma_{y\in U}(X\cap \widetilde{N}_{x}^{\gamma})(y)}{\Sigma_{y\in U}\widetilde{N}_{x}^{\gamma}(y)}\geq\beta$. So $\Sigma_{y\in U}(X\cap \widetilde{N}_{x}^{\gamma})(y)\geq\beta\Sigma_{y\in U}\widetilde{N}_{x}^{\gamma}(y)$. Therefore, $\overline{FR}_{(\alpha,\beta)}(X)=\{x\in U\mid  \Sigma_{y\in U}(X\cap \widetilde{N}_{x}^{\gamma})(y)\geq\beta\Sigma_{y\in U}\widetilde{N}_{x}^{\gamma}(y)\}.$

By Definition 4.3, we have $\underline{FR}_{(\alpha,\beta)}(X)=\{x\in U\mid P(X|\widetilde{N}_{x}^{\gamma})\geq\alpha \}$, which implies that $P(X|\widetilde{N}_{x}^{\gamma})\geq\alpha$ for $x\in \underline{FR}_{(\alpha,\beta)}(X)$. It follows that $P(X|\widetilde{N}_{x}^{\gamma})=\frac{\Sigma_{y\in U}(X\cap \widetilde{N}_{x}^{\gamma})(y)}{\Sigma_{y\in U}\widetilde{N}_{x}^{\gamma}(y)}\geq\alpha$. So $\Sigma_{y\in U}(X\cap \widetilde{N}_{x}^{\gamma})(y)\geq\alpha\Sigma_{y\in U}\widetilde{N}_{x}^{\gamma}(y)$, which implies $\Sigma_{y\in U}\widetilde{N}_{x}^{\gamma}(y)-\Sigma_{y\in U}(X\cap \widetilde{N}_{x}^{\gamma})(y)\leq\Sigma_{y\in U}\widetilde{N}_{x}^{\gamma}(y)-\alpha\Sigma_{y\in U}\widetilde{N}_{x}^{\gamma}(y)$. Therefore, $\underline{FR}_{(\alpha,\beta)}(X)=\{x\in U\mid  \Sigma_{y\in U}(X\cap \widetilde{N}_{x}^{\gamma})(y)\geq\beta\Sigma_{y\in U}\widetilde{N}_{x}^{\gamma}(y)\}.$
$\Box$

\begin{theorem}
Let $(U,\mathscr{C})$ be a fuzzy $\gamma-$covering approximation space, where $U=\{x_{1},x_{2},...,x_{n}\}$, $\mathscr{C}=\{C_{1},C_{2},...,C_{m}\}$, $X\in \mathscr{F}(U)$, $\gamma\in (0,1]$, and $k\in R$. Then
\begin{eqnarray*}
\overline{GR}_{k}(X)&=&\{x\in U\mid P(X|\widetilde{N}_{x}^{\gamma})\geq\frac{k}{\Sigma_{y\in U} \widetilde{N}_{x}^{\gamma}(y)}\};\\
\underline{GR}_{k}(X)&=&\{x\in U\mid P(X|\widetilde{N}_{x}^{\gamma})\geq 1-\frac{k}{\Sigma_{y\in U} \widetilde{N}_{x}^{\gamma}(y)}\}.
\end{eqnarray*}
\end{theorem}

\noindent\textbf{Proof.} By Definition 4.8, we have $\overline{GR}_{k}(X)=\{x\in U\mid \Sigma_{y\in U}(X\cap \widetilde{N}_{x}^{\gamma})(y)\geq k\},$ which implies that $\Sigma_{y\in U}(X\cap \widetilde{N}_{x}^{\gamma})(y)>k$ for $x\in \overline{GR}_{k}(X)$. It follows that $P(X|\widetilde{N}_{x}^{\gamma})=\frac{\Sigma_{y\in U}(X\cap \widetilde{N}_{x}^{\gamma})(y)}{\Sigma_{y\in U}\widetilde{N}_{x}^{\gamma}(y)}\geq\frac{k}{\Sigma_{y\in U}\widetilde{N}_{x}^{\gamma}(y)}$ for $x\in \overline{GR}_{k}(X)$. Therefore, $\overline{GR}_{k}(X)=\{x\in U\mid P(X|\widetilde{N}_{x}^{\gamma})\geq\frac{k}{\Sigma_{y\in U} \widetilde{N}_{x}^{\gamma}(y)}\}.$

By Definition 4.8, we have $\underline{GR}_{k}(X)=\{x\in U\mid \Sigma_{y\in U}[\widetilde{N}_{x}^{\gamma}(y)-(X\cap \widetilde{N}_{x}^{\gamma})(y)]\leq k\},$ which implies that $\Sigma_{y\in U}(X\cap \widetilde{N}_{x}^{\gamma})(y)\geq \Sigma_{y\in U}\widetilde{N}_{x}^{\gamma}(y)-k$ for $x\in \underline{GR}_{k}(X)$. It follows that $P(X|\widetilde{N}_{x}^{\gamma})=\frac{\Sigma_{y\in U}(X\cap \widetilde{N}_{x}^{\gamma})(y)}{\Sigma_{y\in U}\widetilde{N}_{x}^{\gamma}(y)}>1-\frac{k}{\Sigma_{y\in U}\widetilde{N}_{x}^{\gamma}(y)}$ for $x\in \underline{GR}_{k}(X)$. Therefore, $\underline{GR}_{k}(X)=\{x\in U\mid P(X|\widetilde{N}_{x}^{\gamma})\geq 1-\frac{k}{\Sigma_{y\in U} \widetilde{N}_{x}^{\gamma}(y)}\}.$

Theorems 4.13 and 4.14 illustrate the relationship between the fuzzy $\gamma-$covering based probabilistic lower and upper approximations of fuzzy sets and the fuzzy $\gamma-$covering based grade lower and upper approximations of fuzzy sets, which build a bridge between two fuzzy $\gamma-$covering based approximation operators.

\section{Double-quantitative lower and upper approximation operators}

In this section, we present the fuzzy $\gamma-$covering based double-quantitative lower and upper approximations of fuzzy sets in the fuzzy $\gamma-$covering approximation space.

\begin{definition}
Let $(U,\mathscr{C})$ be a fuzzy $\gamma-$covering approximation space, where $U=\{x_{1},x_{2},...,x_{n}\}$, $\mathscr{C}=\{C_{1},C_{2},...,C_{m}\}$, $0\leq \beta \leq \alpha \leq 1,$ and $k\in R$. Then the fuzzy $\gamma-$covering based disjunctive double-quantitative lower and upper approximations of $X\in \mathscr{F}(U)$ are defined as follows:
\begin{eqnarray*}
\overline{DR}_{(\alpha,\beta)\wedge k}^{I}(X)&=&\{x\in U\mid [P(X|\widetilde{N}_{x}^{\gamma})\geq\beta] \wedge [\Sigma_{y\in U}(X\cap \widetilde{N}_{x}^{\gamma})(y)>k]\};\\
\underline{DR}_{(\alpha,\beta)\wedge k}^{I}(X)&=&\{x\in U\mid [P(X|\widetilde{N}_{x}^{\gamma})\geq\alpha] \wedge [\Sigma_{y\in U}[\widetilde{N}_{x}^{\gamma})(y)-(X\cap \widetilde{N}_{x}^{\gamma})(y)]\leq k]\}.
\end{eqnarray*}
\end{definition}

The fuzzy $\gamma-$covering based disjunctive lower and upper approximation operators $\overline{DR}_{(\alpha,\beta)\wedge k}^{I}(X)$
and $\underline{DR}_{(\alpha,\beta)\wedge k}^{I}(X)$ consider the relative and absolute quantitative information, which are generalizations of disjunctive double quantitative rough set model proposed by Xu\cite{Xu}.

We convert the fuzzy $\gamma-$covering based disjunctive double-quantitative lower and upper approximations of the fuzzy set $X\in \mathscr{F}(U)$ as follows:
\begin{eqnarray*}
\overline{DR}_{(\alpha,\beta)\wedge k}^{I}(X)&=&\{x\in U\mid P(X|\widetilde{N}_{x}^{\gamma})\geq max\{\beta,\frac{k}{\Sigma_{y\in U} \widetilde{N}_{x}^{\gamma}(y)}\}\};\\
\underline{DR}_{(\alpha,\beta)\wedge k}^{I}(X)&=&\{x\in U\mid P(X|\widetilde{N}_{x}^{\gamma})\geq max\{\alpha,1-\frac{k}{\Sigma_{y\in U} \widetilde{N}_{x}^{\gamma}(y)}\},
\end{eqnarray*}
and
\begin{eqnarray*}
\overline{DR}_{(\alpha,\beta)\wedge k}^{I}(X)&=&\{x\in U\mid \Sigma_{y\in U}(X\cap \widetilde{N}_{x}^{\gamma})(y)\geq max\{\beta\Sigma_{y\in U}\widetilde{N}_{x}^{\gamma}(y), k\}\};\\
\underline{DR}_{(\alpha,\beta)\wedge k}^{I}(X)&=&\{x\in U\mid \Sigma_{y\in U}(X\cap \widetilde{N}_{x}^{\gamma})(y)\geq max\{\alpha\Sigma_{y\in U}\widetilde{N}_{x}^{\gamma}(y), \Sigma_{y\in U}\widetilde{N}_{x}^{\gamma}(y)-k\}\}.
\end{eqnarray*}

We employ an example to illustrate the computing of the fuzzy $\gamma-$covering based disjunctive double-quantitative lower and upper approximations of sets as follows.

\begin{example}
\begin{upshape}
(Continuation from Example 4.4) Taking $\alpha=0.75,$ $\beta=0.25$, and $k=2$, we have the fuzzy $\gamma-$covering based disjunctive double-quantitative lower and upper approximations of the fuzzy set $X$ as follows:
\begin{eqnarray*}
\underline{DR}_{(\alpha,\beta)\wedge k}^{I}(X)=\{x_{3},x_{6}\}\text{ and }
\overline{DR}_{(\alpha,\beta)\wedge k}^{I}(X)=\{x_{1},x_{2},x_{3},x_{4},x_{5},x_{6},x_{7},x_{8}\};
\end{eqnarray*}
\end{upshape}
\end{example}

In what follows, we show the relationship between  the fuzzy $\gamma-$covering based disjunctive double-quantitative lower and upper approximation operators and the fuzzy $\gamma-$covering based probabilistic and grade lower and upper approximation operators.

\begin{theorem}
Let $(U,\mathscr{C})$ be a fuzzy $\gamma-$covering approximation space, where $U=\{x_{1},x_{2},...,x_{n}\}$, $\mathscr{C}=\{C_{1},C_{2},...,C_{m}\}$, $0\leq \beta \leq \alpha \leq 1,$ $k\in R$, and $X\in \mathscr{F}(U)$. Then
\begin{eqnarray*}
\overline{DR}_{(\alpha,\beta)\wedge k}^{I}(X)&=&\overline{FR}_{(\alpha,\beta)}(X)\cap \overline{GR}_{ k}(X);\\
\underline{DR}_{(\alpha,\beta)\wedge k}^{I}(X)&=&\underline{FR}_{(\alpha,\beta)}(X)\cap \underline{GR}_{ k}(X).
\end{eqnarray*}
\end{theorem}

\noindent\textbf{Proof.} For $z\in\overline{DR}_{(\alpha,\beta)\wedge k}^{I}(X)$, we have
$P(X|\widetilde{N}_{z}^{\gamma})\geq\beta $ and $\Sigma_{y\in U}(X\cap \widetilde{N}_{z}^{\gamma})(y)>k$, which implies that $z\in\overline{FR}_{(\alpha,\beta)}(X)$ and $ z\in\overline{GR}_{k}(X)$. It follows that $\overline{DR}_{(\alpha,\beta)\wedge k}^{I}(X)\subseteq\overline{FR}_{(\alpha,\beta)}(X)\cap \overline{GR}_{k}(X).$ For $z\in\overline{FR}_{(\alpha,\beta)}(X)\cup \overline{GR}_{k}(X)$, we have $P(X|\widetilde{N}_{z}^{\gamma})\geq\beta $ and $\Sigma_{y\in U}(X\cap \widetilde{N}_{z}^{\gamma})(y)\geq k$, which implies that $z\in\overline{DR}_{(\alpha,\beta)\wedge k}^{I}(X)$. Therefore, $\overline{DR}_{(\alpha,\beta)\wedge k}^{I}(X)=\overline{FR}_{(\alpha,\beta)}(X)\cap \overline{GR}_{k}(X).$

For $z\in\underline{DR}_{(\alpha,\beta)\wedge k}^{I}(X)$, we have
$P(X|\widetilde{N}_{z}^{\gamma})\geq\alpha $ and $\Sigma_{y\in U}[\widetilde{N}_{x}^{\gamma})(y)-(X\cap \widetilde{N}_{x}^{\gamma})(y)]\leq k$, which implies that $z\in\underline{FR}_{(\alpha,\beta)}(X)$ and $ z\in\underline{GR}_{ k}(X)$. It follows that $\underline{DR}_{(\alpha,\beta)\wedge k}^{I}(X)\subseteq\underline{FR}_{(\alpha,\beta)}(X)\cap \underline{GR}_{k}(X).$  For $z\in \underline{FR}_{(\alpha,\beta)}(X)\cap \underline{GR}_{k}(X)$, we have $P(X|\widetilde{N}_{z}^{\gamma})\geq\alpha $ and $\Sigma_{y\in U}[\widetilde{N}_{x}^{\gamma})(y)-(X\cap \widetilde{N}_{x}^{\gamma})(y)]\leq k$, which implies that $z\in \underline{DR}_{(\alpha,\beta)\wedge k}^{I}(X).$  Therefore, $\underline{DR}_{(\alpha,\beta)\wedge k}^{I}(X)=\underline{FR}_{(\alpha,\beta)}(X)\cap \underline{GR}_{k}(X)$. $\Box$

\begin{example}(Continuation from Example 4.4) Taking $\alpha=0.75,$ $\beta=0.25$, and $k=2$, we have the fuzzy $\gamma-$covering based disjunctive double-quantitative lower and upper approximations of the fuzzy set $X$ as follows:
\begin{eqnarray*}
\underline{DR}_{(\alpha,\beta)\wedge k}^{I}(X)&=&\underline{FR}_{(\alpha,\beta)}(X)\cap\underline{GR}_{2}(X)=\{x_{3},x_{6}\};\\
\overline{DR}_{(\alpha,\beta)\wedge k}^{I}(X)&=&\overline{FR}_{(\alpha,\beta)}(X)\cap\overline{GR}_{2}(X)=\{x_{1},x_{2},x_{3},
x_{4},x_{5},x_{6},x_{7},x_{8}\}.
\end{eqnarray*}
\end{example}

We investigate the basic properties of the fuzzy $\gamma-$covering based disjunctive double-quantitative lower and upper approximation operators as follows.

\begin{theorem}
Let $(U,\mathscr{C})$ be a fuzzy $\gamma-$covering approximation space, where $U=\{x_{1},x_{2},...,x_{n}\}$, $\mathscr{C}=\{C_{1},C_{2},...,C_{m}\}$, $0\leq \beta< \alpha\leq 1$, $k\in R$, and $X,Y\in \mathscr{F}(U)$. Then\\
$(1)\underline{DR}_{(\alpha,\beta)\wedge k}^{I}(U)=U;\overline{DR}_{(\alpha,\beta)\wedge k}^{I}(\emptyset)= \emptyset;\\
(2) X\subseteq Y\Rightarrow\overline{DR}_{(\alpha,\beta)\wedge k}^{I}(X)\subseteq \overline{DR}_{(\alpha,\beta)\wedge k}^{I}(Y);\\
(3) X\subseteq Y\underline{DR}_{(\alpha,\beta)\wedge k}^{I}(X)\subseteq \underline{DR}_{(\alpha,\beta)\wedge k}^{I}(Y);\\
(4) \overline{DR}_{(\alpha,\beta)\wedge k}^{I}(X)\cup \overline{DR}_{(\alpha,\beta)\wedge k}^{I}(Y)\subseteq \overline{DR}_{(\alpha,\beta)\wedge k}^{I}(X\cup Y);\\
(5)\underline{DR}_{(\alpha,\beta)\wedge k}^{I}(X)\cup \underline{DR_{(\alpha,\beta)\wedge k}}^{I}(Y)\subseteq \underline{DR}_{(\alpha,\beta)\wedge k}^{I}(X\cup Y);\\
(6) \overline{DR}_{(\alpha,\beta)\wedge k}^{I}(X\cap Y)\subseteq \overline{DR}_{(\alpha,\beta)\wedge k}^{I}(X)\cap \overline{DR}_{(\alpha,\beta)\wedge k}^{I}(Y);\\
 (7) \underline{DR}_{(\alpha,\beta)\wedge k}^{I}(X\cap Y)\subseteq \underline{DR}_{(\alpha,\beta)\wedge k}^{I}(X)\cap \underline{DR}_{(\alpha,\beta)\wedge k}^{I}(Y);\\
(8) \alpha_{1} \leq \alpha_{2}, \beta_{1} \leq \beta_{2}, k_{1} \leq k_{2}\Rightarrow\underline{DR}_{(\alpha_{1},\beta_{1})\wedge k_{1}}^{I}(X)\subseteq \underline{DR}_{(\alpha_{2},\beta_{2})\wedge k_{2}}^{I}(X);\\
(9) \alpha_{1} \leq \alpha_{2}, \beta_{1} \leq \beta_{2}, k_{1} \leq k_{2}\Rightarrow\overline{DR}_{(\alpha_{1},\beta_{1})\wedge k_{1}}(X)\subseteq \overline{DR}_{(\alpha_{2},\beta_{2})\wedge k_{2}}^{I}(X).$
\end{theorem}

\noindent\textbf{Proof.}
By Theorems 4.7 and 4.12, the proof is straightforward.
$\Box$

\begin{definition}
Let $(U,\mathscr{C})$ be a fuzzy $\gamma-$covering approximation space, where $U=\{x_{1},x_{2},...,x_{n}\}$, $\mathscr{C}=\{C_{1},C_{2},...,C_{m}\}$, $0\leq \beta \leq \alpha \leq 1,$ $k\in R$, and $X\in \mathscr{F}(U)$. Then the fuzzy $\gamma-$covering based conjunctive double-quantitative lower and upper approximations of the fuzzy set $X\in \mathscr{F}(U)$ are defined as follows:
\begin{eqnarray*}
\overline{DR}_{(\alpha,\beta)\vee k}^{II}(X)&=&\{x\in U\mid [P(X|\widetilde{N}_{x}^{\gamma})\geq\beta] \vee [\Sigma_{y\in U}(X\cap \widetilde{N}_{x}^{\gamma})(y)>k]\};\\
\underline{DR}_{(\alpha,\beta)\vee k}^{II}(X)&=&\{x\in U\mid [P(X|\widetilde{N}_{x}^{\gamma})\geq\alpha] \vee [\Sigma_{y\in U}(X^{c}\cap \widetilde{N}_{x}^{\gamma})(y)\leq k]\}.
\end{eqnarray*}
\end{definition}

The fuzzy $\gamma-$covering based conjunctive double-quantitative lower and upper approximation operators $\overline{DR}_{(\alpha,\beta)\wedge k}^{II}(X)$
and $\underline{DR}_{(\alpha,\beta)\wedge k}^{II}(X)$ for the fuzzy set $X\in \mathscr{F}(U)$ consider the relative and absolute quantitative information, which are generalizations of conjunctive double-quantitative rough set model proposed by Xu\cite{Xu}.

We convert the fuzzy $\gamma-$covering based disjunctive double-quantitative lower and upper approximations of the fuzzy set $X\in \mathscr{F}(U)$ as follows:
\begin{eqnarray*}
\overline{DR}_{(\alpha,\beta)\vee k}^{II}(X)&=&\{x\in U\mid P(X|\widetilde{N}_{x}^{\gamma})\geq min\{\beta,\frac{k}{\Sigma_{y\in U} \widetilde{N}_{x}^{\gamma}(y)}\}\};\\
\underline{DR}_{(\alpha,\beta)\vee k}^{II}(X)&=&\{x\in U\mid P(X|\widetilde{N}_{x}^{\gamma})\geq min\{\alpha,1-\frac{k}{\Sigma_{y\in U} \widetilde{N}_{x}^{\gamma}(y)}\},
\end{eqnarray*}
and
\begin{eqnarray*}
\overline{DR}_{(\alpha,\beta)\vee k}^{II}(X)&=&\{x\in U\mid \Sigma_{y\in U}(X\cap \widetilde{N}_{x}^{\gamma})(y)\geq min\{\beta\Sigma_{y\in U}\widetilde{N}_{x}^{\gamma}(y), k\}\};\\
\underline{DR}_{(\alpha,\beta)\vee k}^{II}(X)&=&\{x\in U\mid \Sigma_{y\in U}(X\cap \widetilde{N}_{x}^{\gamma})(y)\geq min\{\alpha\Sigma_{y\in U}\widetilde{N}_{x}^{\gamma}(y), \Sigma_{y\in U}\widetilde{N}_{x}^{\gamma}(y)-k\}\}.
\end{eqnarray*}

We employ an example to illustrate the computing of the fuzzy $\gamma-$covering based conjunctive double-quantitative lower and upper approximations of fuzzy sets as follows.

\begin{example}(Continuation from Example 4.4) Taking $\alpha=0.75,$ $\beta=0.25$, and $k=2$, we have the conjunctive double-quantitative lower and upper approximations of the fuzzy set $X$ as follows:
\begin{eqnarray*}
\underline{DR}_{(\alpha,\beta)\vee k}^{II}(X)=\{x_{2},x_{3},x_{6},x_{8}\}\text{ and }
\overline{DR}_{(\alpha,\beta)\vee k}^{II}(X)=\{x_{1},x_{2},x_{3},x_{4},x_{5},x_{6},x_{7},x_{8}\}.
\end{eqnarray*}
\end{example}

We show the relationship between the fuzzy $\gamma-$covering based conjunctive double-quantitative lower and upper approximation operators and the fuzzy $\gamma-$covering based probabilistic and grade lower and upper approximation operators as follows.

\begin{theorem}
Let $(U,\mathscr{C})$ be a fuzzy $\gamma-$covering approximation space, where $U=\{x_{1},x_{2},...,x_{n}\}$, $\mathscr{C}=\{C_{1},C_{2},...,C_{m}\}$, $0\leq \beta \leq \alpha \leq 1,$ $X\in \mathscr{F}(U)$, and $k\in R$. Then
\begin{eqnarray*}
\overline{DR}_{(\alpha,\beta)\vee k}^{II}(X)&=&\overline{FR}_{(\alpha,\beta)}(X)\cup \overline{GR}_{ k}(X);\\
\underline{DR}_{(\alpha,\beta)\vee k}^{II}(X)&=&\underline{FR}_{(\alpha,\beta)}(X)\cup \underline{GR}_{ k}(X).
\end{eqnarray*}
\end{theorem}

\noindent\textbf{Proof.} For $z\in\overline{DR}_{(\alpha,\beta)\wedge k}^{II}(X)$, we have
$P(X|\widetilde{N}_{z}^{\gamma})\geq\beta $ or $\Sigma_{y\in U}(X\cap \widetilde{N}_{z}^{\gamma})(y)\geq k$, which implies that $z\in\overline{FR}_{(\alpha,\beta)}(X)$ or $ z\in\overline{GR}_{ k}(X)$. It follows that $\overline{DR}_{(\alpha,\beta)\wedge k}^{II}(X)\subseteq\overline{FR}_{(\alpha,\beta)}(X)\cup \overline{GR}_{ k}(X).$ For $z\in\overline{FR}_{(\alpha,\beta)}(X)\cup \overline{GR}_{k}(X)$, we have $P(X|\widetilde{N}_{z}^{\gamma})\geq\beta $ or $\Sigma_{y\in U}(X\cap \widetilde{N}_{z}^{\gamma})(y)\geq k$, which implies that $z\in\overline{DR}_{(\alpha,\beta)\vee k}^{II}(X)$. Therefore, $\overline{DR}_{(\alpha,\beta)\vee k}^{II}(X)=\overline{FR}_{(\alpha,\beta)}(X)\cup \overline{GR}_{ k}(X).$

For $z\in\underline{DR}_{(\alpha,\beta)\wedge k}^{II}(X)$, we have
$P(X|\widetilde{N}_{z}^{\gamma})\geq\alpha $ or $\Sigma_{y\in U}[\widetilde{N}_{x}^{\gamma})(y)-(X\cap \widetilde{N}_{x}^{\gamma})(y)]\leq k$, which implies that $z\in\underline{FR}_{(\alpha,\beta)}(X)$ or $ z\in\underline{GR}_{ k}(X)$. It follows that $\underline{DR}_{(\alpha,\beta)\wedge k}^{II}(X)\subseteq\underline{FR}_{(\alpha,\beta)}(X)\cup \underline{GR}_{ k}(X).$  For $z\in \underline{FR}_{(\alpha,\beta)}(X)\cup \underline{GR}_{ k}(X)$, we have $P(X|\widetilde{N}_{z}^{\gamma})\geq\alpha $ or $\Sigma_{y\in U}[\widetilde{N}_{x}^{\gamma})(y)-(X\cap \widetilde{N}_{x}^{\gamma})(y)]\leq k$, which implies that $z\in \underline{DR}_{(\alpha,\beta)\vee k}^{II}(X).$  Therefore, $\underline{DR}_{(\alpha,\beta)\vee k}^{II}(X)=\underline{FR}_{(\alpha,\beta)}(X)\cup \underline{GR}_{k}(X)$. $\Box$

\begin{example}
(Continuation from Example 4.4) Taking $\alpha=0.75,$ $\beta=0.25$, and $k=2$, we have the fuzzy $\gamma-$covering based conjunctive double-quantitative lower and upper approximations of the fuzzy set $X$ as follows:
\begin{eqnarray*}
\underline{DR}_{(\alpha,\beta)\vee k}^{II}(X)&=&\underline{FR}_{(\alpha,\beta)}(X)\cup\underline{GR}_{2}(X)=\{x_{2},x_{3},x_{6},x_{8}\};\\
\overline{DR}_{(\alpha,\beta)\vee k}^{II}(X)&=&\overline{FR}_{(\alpha,\beta)}(X)\cup\overline{GR}_{2}(X)=\{x_{1},x_{2},x_{3},x_{4},
x_{5},x_{6},x_{7},x_{8}\}.
\end{eqnarray*}
\end{example}

We show the basic properties of the fuzzy $\gamma-$covering based conjunctive double-quantitative lower and upper approximation operators as follows.

\begin{theorem}
Let $(U,\mathscr{C})$ be a fuzzy $\gamma-$covering approximation space, where $U=\{x_{1},x_{2},...,x_{n}\}$, $\mathscr{C}=\{C_{1},C_{2},...,C_{m}\}$, $0\leq \beta< \alpha\leq 1$, $k\in R$, and $X,Y\in \mathscr{F}(U)$. Then\\
$(1)\underline{DR}_{(\alpha,\beta)\vee k}^{II}(U)=U;\overline{DR}_{(\alpha,\beta)\vee k}^{II}(\emptyset)= \emptyset;\\
(2) X\subseteq Y\Rightarrow\overline{DR}_{(\alpha,\beta)\vee k}^{II}(X)\subseteq \overline{DR}_{(\alpha,\beta)\vee k}^{II}(Y);\\
(3) X\subseteq Y\Rightarrow\underline{DR}_{(\alpha,\beta)\vee k}^{II}(X)\subseteq \underline{DR}_{(\alpha,\beta)\vee k}^{II}(Y);\\
(4) \overline{DR}_{(\alpha,\beta)\vee k}^{II}(X)\cup \overline{DR}_{(\alpha,\beta)\vee k}^{II}(Y)\subseteq \overline{DR}_{(\alpha,\beta)\vee k}^{II}(X\cup Y);\\
(5)\underline{DR}_{(\alpha,\beta)\vee k}^{II}(X)\cup \underline{DR_{(\alpha,\beta)\vee k}}^{II}(Y)\subseteq \underline{DR}_{(\alpha,\beta)\vee k}^{II}(X\cup Y);\\
(6) \overline{DR}_{(\alpha,\beta)\vee k}^{II}(X\cap Y)\subseteq \overline{DR}_{(\alpha,\beta)\vee k}^{II}(X)\cap \overline{DR}_{(\alpha,\beta)\vee k}^{II}(Y);\\
(7) \underline{DR}_{(\alpha,\beta)\vee k}^{II}(X\cap Y)\subseteq \underline{DR}_{(\alpha,\beta)\vee k}^{II}(X)\cap \underline{DR}_{(\alpha,\beta)\vee k}^{II}(Y);\\
(8) \alpha_{1} \leq \alpha_{2}, \beta_{1} \leq \beta_{2}, k_{1} \leq k_{2}\Rightarrow\underline{DR}_{(\alpha_{1},\beta_{1})\vee k_{1}}^{II}(X)\subseteq \underline{DR}_{(\alpha_{2},\beta_{2})\vee k_{2}}^{II}(X);\\
(9) \alpha_{1} \leq \alpha_{2}, \beta_{1} \leq \beta_{2}, k_{1} \leq k_{2}\Rightarrow\overline{DR}_{(\alpha_{1},\beta_{1})\vee k_{1}}(X)\subseteq \overline{DR}_{(\alpha_{2},\beta_{2})\vee k_{2}}^{II}(X).$
\end{theorem}

\noindent\textbf{Proof.}
By Theorems 4.7 and 4.12, the proof is straightforward.
$\Box$

\section{Multi-granulation double-quantitative approximation operators}

In this section, we present the fuzzy $\gamma^{\ast}-$coverings based multi-granulation double-quantitative lower and upper approximation operators in the fuzzy $\gamma^{\ast}-$coverings information system.

\subsection{Multi-granulation probabilistic approximation operators}

In this section, we propose the concept of the fuzzy $\gamma^{\ast}-$coverings information system.

\begin{definition}
Let $U$ be a finite universe, $\Delta^{\ast}$ a family of fuzzy
$\gamma-$coverings, where $U=\{x_{1},x_{2},...,x_{n}\}$, $\Delta^{\ast}=\{\mathscr{C}_{1},
\mathscr{C}_{2},...,\mathscr{C}_{m}\}$, and $\mathscr{C}_{i}$ a fuzzy
$\gamma_{i}-$covering of $U$. Then $(U,\Delta^{\ast})$ is called a
fuzzy $\gamma^{\ast}-$coverings information system, where $\gamma^{\ast}=[\gamma_{1},\gamma_{2},...,\gamma_{|\Delta^{\ast}|}]$.
\end{definition}

The fuzzy $\gamma^{\ast}-$coverings information system is the generalization of the covering information system, and we also can view the fuzzy $\gamma^{\ast}-$coverings information system as a $\gamma-$covering approximation space, where $\gamma=min\{\gamma_{i}|1\leq i\leq m\}$.

\begin{example}Let $U=\{x_{1},x_{2},x_{3},x_{4},x_{5},x_{6},x_{7},x_{8}\}$, $\triangle^{\ast}=\{\mathscr{C}^{\ast}_{1},
\mathscr{C}^{\ast}_{2}\}$, $\mathscr{C}^{\ast}_{1}=\{C_{11},C_{12},C_{13}\}$, and $\mathscr{C}^{\ast}_{2}=\{C_{21},C_{22},C_{23}\}$, where $\gamma^{\ast}=[0.9,0.6],$ and
\begin{eqnarray*}
C_{11}&=&\frac{1}{x_{1}}+\frac{0.7}{x_{2}}+\frac{0}{x_{3}}
+\frac{0.9}{x_{4}}+\frac{0.9}{x_{5}}
+\frac{0}{x_{6}}+\frac{0.9}{x_{7}}+\frac{0.8}{x_{8}};\\
C_{12}&=&\frac{0.6}{x_{1}}+\frac{0.9}{x_{2}}+\frac{0.4}{x_{3}}
+\frac{0.4}{x_{4}}+\frac{0.5}{x_{5}}
+\frac{0.7}{x_{6}}+\frac{0.5}{x_{7}}+\frac{1}{x_{8}};\\
C_{13}&=&\frac{0}{x_{1}}+\frac{0.5}{x_{2}}+\frac{0.9}{x_{3}}
+\frac{0}{x_{4}}+\frac{0.5}{x_{5}}
+\frac{0.9}{x_{6}}+\frac{0}{x_{7}}+\frac{0.5}{x_{8}};
\\
C_{21}&=&\frac{0.6}{x_{1}}+\frac{0.4}{x_{2}}+\frac{0.2}{x_{3}}
+\frac{0.4}{x_{4}}+\frac{0.1}{x_{5}}
+\frac{0.6}{x_{6}}+\frac{0.6}{x_{7}}+\frac{0.5}{x_{8}};\\
C_{22}&=&\frac{0.5}{x_{1}}+\frac{0.3}{x_{2}}+\frac{0.6}{x_{3}}
+\frac{0.6}{x_{4}}+\frac{0.4}{x_{5}}
+\frac{0.5}{x_{6}}+\frac{0.2}{x_{7}}+\frac{0.6}{x_{8}};\\
C_{23}&=&\frac{0.2}{x_{1}}+\frac{0.6}{x_{2}}+\frac{0.2}{x_{3}}
+\frac{0.5}{x_{4}}+\frac{0.6}{x_{5}}
+\frac{0.3}{x_{6}}+\frac{0}{x_{7}}+\frac{0.3}{x_{8}}.
\end{eqnarray*}
Then we have a
fuzzy $\gamma^{\ast}-$coverings information system $(U,\Delta^{\ast})$.
\end{example}

In what follows, we present the fuzzy $\gamma^{\ast}-$coverings disjunctive
multi-granulation probabilistic lower and upper approximation operators in the fuzzy $\gamma^{\ast}-$coverings information system.

\begin{definition}
Let $(U,\Delta)$ be a fuzzy $\gamma^{\ast}-$coverings information system, and
$0\leq \beta_{i} \leq \alpha_{i} \leq 1$. Then the fuzzy $\gamma^{\ast}-$coverings based disjunctive multi-granulation probabilistic lower and upper approximations of the fuzzy set $X\in\mathscr{F}(U)$ are defined as follows:
\begin{eqnarray*}
\overline{MR}^{I}_{(\alpha^{\ast},\beta^{\ast})}(X)&=&\{x\in U\mid \bigwedge_{1\leq i\leq m}[P(X|\widetilde{N}_{x}^{\gamma_{i}})\geq\beta_{i}]\};\\
\underline{MR}^{I}_{(\alpha^{\ast},\beta^{\ast})}(X)&=&\{x\in U\mid \bigwedge_{1\leq i\leq m}[P(X|\widetilde{N}_{x}^{\gamma_{i}})\geq\alpha_{i}]\}.
\end{eqnarray*}
\end{definition}

The fuzzy $\gamma^{\ast}-$coverings based disjunctive multi-granulation probabilistic lower and upper approximations operators $\overline{MR}^{I}_{(\alpha^{\ast},\beta^{\ast})}(X)$
and $\underline{MR}^{I}_{(\alpha^{\ast},\beta^{\ast})}(X)$ of the fuzzy set $X\in\mathscr{F}(U)$ consider the relative quantitative information, which are generalizations of the disjunctive multi-granulation rough set model proposed by Qian\cite{Qian3}.

We employ an example to illustrate the construction of the fuzzy $\gamma^{\ast}-$coverings based disjunctive multi-granulation probabilistic lower and upper approximations as follows.

\begin{example}(Continuation of Example 4.4) Taking $\alpha_{1}=\alpha_{2}=0.75,$ and $\beta_{1}=\beta_{2}=0.25,$ we have
\begin{eqnarray*}
\underline{MR}^{I}_{(\alpha^{\ast},\beta^{\ast})}(X)=\{x_{3},x_{6}\}\text{ and }
\overline{MR}^{I}_{(\alpha^{\ast},\beta^{\ast})}(X)=\{x_{1},x_{2},x_{3},x_{4},x_{5},x_{6},x_{7},x_{8}\}.
\end{eqnarray*}
\end{example}

We employ the following theorem to illustrate the relationship between the fuzzy $\gamma^{\ast}-$coverings based disjunctive multi-granulation probabilistic approximations operators and the fuzzy $\gamma^{\ast}-$coverings based probabilistic approximations operators.

\begin{theorem}
Let $(U,\Delta)$ be a fuzzy $\gamma^{\ast}-$covering information system, $X\in\mathscr{F}(U)$, and
$0\leq \beta_{i} \leq \alpha_{i} \leq 1, 1\leq i\leq m$. Then we have
\begin{eqnarray*}
 \overline{MR}^{I}_{(\alpha^{\ast},\beta^{\ast})}(X)&=&\bigcap_{1\leq i\leq m}\overline{FR}_{(\alpha_{i},\beta_{i})}(X);\\
\underline{MR}^{I}_{(\alpha^{\ast},\beta^{\ast})}(X)&=&\bigcap_{1\leq i\leq m}\underline{FR}_{(\alpha_{i},\beta_{i})}(X).
\end{eqnarray*}
\end{theorem}

\noindent\textbf{Proof.}
The proof is straightforward by Definition 6.3.
$\Box$

\begin{example}(Continuation of Example 4.4) Taking $\alpha_{1}=\alpha_{2}=0.75,$ and $\beta_{1}=\beta_{2}=0.25,$ we have
\begin{eqnarray*}
\underline{MR}^{I}_{(\alpha^{\ast},\beta^{\ast})}(X)&=&\underline{FR}_{(\alpha_{1},\beta_{1})}(X)\cap \underline{FR}_{(\alpha_{2},\beta_{2})}(X)\\
&=&\{x_{3},x_{6}\}\cap \{x_{1},x_{2},x_{3},x_{4},x_{5},x_{6},x_{7},x_{8}\}\\
&=&\{x_{3},x_{6}\};\\
\overline{MR}^{I}_{(\alpha^{\ast},\beta^{\ast})}(X)
&=&\overline{FR}_{(\alpha_{1},\beta_{1})}(X)\cap \overline{FR}_{(\alpha_{2},\beta_{2})}(X)\\
&=&\{x_{1},x_{2},x_{3},x_{4},x_{5},x_{6},x_{7},x_{8}\}\cap
\{x_{1},x_{2},x_{3},x_{4},x_{5},x_{6},x_{7},x_{8}\}\\
&=&\{x_{1},x_{2},x_{3},x_{4},x_{5},x_{6},x_{7},x_{8}\}.
\end{eqnarray*}
\end{example}

We present the following concept to illustrate the relationship between the fuzzy $\gamma^{\ast}-$coverings based disjunctive multi-granulation probabilistic lower and upper approximation operators.

\begin{definition}
Let $\alpha^{\ast}_{1}=[\alpha_{11},\alpha_{12},...,\alpha_{1m}]$, $\alpha^{\ast}_{2}=[\alpha_{21},\alpha_{22},...,\alpha_{2m}]$, $\beta^{\ast}_{1}=[\beta_{11},\beta_{12},...,\beta_{1m}]$ and $\beta^{\ast}_{2}=[\beta_{21},\beta_{22},...,\beta_{2m}]$, where $0\leq\alpha_{ij}, \beta_{ij}\leq 1$. Then we say $(\alpha^{\ast}_{1},\beta^{\ast}_{1})\leq (\alpha^{\ast}_{2},\beta^{\ast}_{2})$ if $\alpha_{1i}\leq \alpha_{2i}$ and $\beta_{1i}\leq \beta_{2i}$ for $1\leq i\leq m$.
\end{definition}

We provide the basic properties of the fuzzy $\gamma^{\ast}-$coverings based disjunctive multi-granulation probabilistic lower and upper approximation operators.

\begin{theorem}
Let $U$ be a finite universe, $\Delta^{\ast}$ a family of fuzzy
$\gamma-$coverings, where
$U=\{x_{1},x_{2},...,x_{n}\}$, $\Delta^{\ast}=\{\mathscr{C}_{1},
\mathscr{C}_{2},...,\mathscr{C}_{m}\}$, $\mathscr{C}_{i}$ a fuzzy
$\gamma_{i}-$covering of $U$, and $X,Y\in\mathscr{F}(U)$. Then\\
$(1)\underline{MR}^{I}_{(\alpha^{\ast},\beta^{\ast})}(U)=U;\overline{MR}^{I}_{(\alpha^{\ast},\beta^{\ast})}(\emptyset)= \emptyset;\\
(2) X\subseteq Y\Rightarrow\overline{MR}^{I}_{(\alpha^{\ast},\beta^{\ast})}(X)\subseteq \overline{MR}^{I}_{(\alpha^{\ast},\beta^{\ast})}(Y);\\
(3) X\subseteq Y\Rightarrow\underline{MR}^{I}_{(\alpha^{\ast},\beta^{\ast})}(X)\subseteq \underline{MR}^{I}_{(\alpha^{\ast},\beta^{\ast})}(Y);\\
(4) \overline{MR}^{I}_{(\alpha^{\ast},\beta^{\ast})}(X)\cup \overline{MR}^{I}_{(\alpha^{\ast},\beta^{\ast})}(X)\subseteq \overline{MR}^{I}_{(\alpha^{\ast},\beta^{\ast})}(X\cup Y);\\
(5) \underline{MR}^{I}_{(\alpha^{\ast},\beta^{\ast})}(X)\cup \underline{MR}^{I}_{(\alpha^{\ast},\beta^{\ast})}(X)\subseteq \underline{MR}^{I}_{(\alpha^{\ast},\beta^{\ast})}(X\cup Y);\\
(6) \overline{MR}^{I}_{(\alpha^{\ast},\beta^{\ast})}(X\cap Y)\subseteq \overline{MR}^{I}_{(\alpha^{\ast},\beta^{\ast})}(X)\cap \overline{MR}^{I}_{(\alpha^{\ast},\beta^{\ast})}(X); \\
(7) \underline{MR}^{I}_{(\alpha^{\ast},\beta^{\ast})}(X\cap Y)\subseteq \underline{MR}^{I}_{(\alpha^{\ast},\beta^{\ast})}(X)\cap \underline{MR}^{I}_{(\alpha^{\ast},\beta^{\ast})}(Y);\\
(8) \alpha^{\ast}_{1} \leq \alpha^{\ast}_{2},\beta^{\ast}_{1} \leq \beta^{\ast}_{2} \Rightarrow\underline{MR}^{I}_{(\alpha^{\ast}_{2},\beta^{\ast}_{2})}(X)\subseteq \underline{MR}^{I}_{(\alpha^{\ast}_{1},\beta^{\ast}_{1})}(Y); \\
(9) \alpha^{\ast}_{1} \leq \alpha^{\ast}_{2},\beta^{\ast}_{1} \leq \beta^{\ast}_{2}  \Rightarrow\overline{MR}^{I}_{(\alpha^{\ast}_{2},\beta^{\ast}_{2})}(X)\subseteq \overline{MR}^{I}_{(\alpha^{\ast}_{1},\beta^{\ast}_{1})}(X).$
\end{theorem}

\noindent\textbf{Proof.}
By Theorem 4.7 and Definition 6.9, the proof is straightforward.
$\Box$

In what follows, we present the fuzzy $\gamma^{\ast}-$coverings based conjunctive
multi-granulation probabilistic lower and upper approximation operators for the fuzzy $\gamma^{\ast}-$coverings information system.

\begin{definition}
Let $(U,\Delta)$ be a fuzzy $\gamma^{\ast}-$coverings information system, and
$0\leq \beta_{i} \leq \alpha_{i} \leq 1$. Then the fuzzy $\gamma^{\ast}-$coverings based conjunctive
multi-granulation probabilistic lower and upper approximations of the fuzzy set $X\in
\mathscr{F}(U)$ are defined as follows:
\begin{eqnarray*}
\overline{MR}^{II}_{(\alpha^{\ast},\beta^{\ast})}(X)&=&\{x\in U\mid \bigvee_{1\leq i\leq m}[P(X|\widetilde{N}_{x}^{\gamma_{i}})\geq\beta_{i}]\};\\
\underline{MR}^{II}_{(\alpha^{\ast},\beta^{\ast})}(X)&=&\{x\in U\mid \bigvee_{1\leq i\leq m}[P(X|\widetilde{N}_{x}^{\gamma_{i}})\geq\alpha_{i}]\}.
\end{eqnarray*}
\end{definition}

The fuzzy $\gamma^{\ast}-$coverings based conjunctive multi-granulation probabilistic lower and upper approximations operators $\overline{MR}^{II}_{(\alpha^{\ast},\beta^{\ast})}(X)$
and $\underline{MR}^{II}_{(\alpha^{\ast},\beta^{\ast})}(X)$ of the fuzzy set $X\in\mathscr{F}(U)$ also consider the relative quantitative information, which are generalizations of the conjunctive multi-granulation rough set model proposed by Qian\cite{Qian3}.

We employ an example to illustrate the construction of the fuzzy $\gamma^{\ast}-$coverings based conjunctive multi-granulation probabilistic lower and upper approximations as follows.

\begin{example}(Continuation of Example 5.2) Taking $\alpha_{1}=\alpha_{2}=0.75,$ and $\beta_{1}=\beta_{2}=0.25,$ we have
\begin{eqnarray*}
\underline{MR}^{II}_{(\alpha^{\ast},\beta^{\ast})}(X)&=&\{x_{1},x_{2},x_{3},x_{4},x_{5},x_{6},x_{7},x_{8}\};\\
\overline{MR}^{II}_{(\alpha^{\ast},\beta^{\ast})}(X)&=&\{x_{1},x_{2},x_{3},x_{4},x_{5},x_{6},x_{7},x_{8}\}.
\end{eqnarray*}
\end{example}

We employ the following theorem to illustrate the relationship between the fuzzy $\gamma^{\ast}-$coverings based conjunctive multi-granulation probabilistic approximations operators and the fuzzy $\gamma^{\ast}-$coverings based probabilistic approximations operators.

\begin{theorem}
Let $(U,\Delta)$ be a fuzzy $\gamma^{\ast}-$covering information system,
$0\leq \beta_{i} \leq \alpha_{i} \leq 1, 1\leq i\leq m$, and $X\in\mathscr{F}(U)$. Then we have
\begin{eqnarray*}
\overline{MR}^{II}_{(\alpha^{\ast},\beta^{\ast})}(X)&=&\bigcup_{1\leq i\leq m}\overline{FR}_{(\alpha_{i},\beta_{i})}(X);\\
\underline{MR}^{II}_{(\alpha^{\ast},\beta^{\ast})}(X)&=&\bigcup_{1\leq i\leq m}\underline{FR}_{(\alpha_{i},\beta_{i})}(X).
\end{eqnarray*}
\end{theorem}

\noindent\textbf{Proof.}
The proof is straightforward by Definition 6.3.
$\Box$

\begin{example}(Continuation of Example 5.4) Taking $\alpha_{1}=\alpha_{2}=0.75,$ and $\beta_{1}=\beta_{2}=0.25,$ we have
\begin{eqnarray*}
\underline{MR}^{II}_{(\alpha^{\ast},\beta^{\ast})}(X)&=&\underline{FR}_{(\alpha_{1},\beta_{1})}(X)\cup \underline{FR}_{(\alpha_{2},\beta_{2})}(X)\\
&=&\{x_{3},x_{6}\}\cup \{x_{1},x_{2},x_{3},x_{4},x_{5},x_{6},x_{7},x_{8}\}\\
&=&\{x_{1},x_{2},x_{3},x_{4},x_{5},x_{6},x_{7},x_{8}\};\\
\overline{MR}^{II}_{(\alpha^{\ast},\beta^{\ast})}(X)
&=&\overline{FR}_{(\alpha_{1},\beta_{1})}(X)\cup \overline{FR}_{(\alpha_{2},\beta_{2})}(X)\\
&=&\{x_{1},x_{2},x_{3},x_{4},x_{5},x_{6},x_{7},x_{8}\}\cup
\{x_{1},x_{2},x_{3},x_{4},x_{5},x_{6},x_{7},x_{8}\}\\
&=&\{x_{1},x_{2},x_{3},x_{4},x_{5},x_{6},x_{7},x_{8}\}.
\end{eqnarray*}
\end{example}

We study the basic properties of the fuzzy $\gamma^{\ast}-$coverings based conjunctive multi-granulation probabilistic lower and upper approximations operators as follows.

\begin{theorem}
Let $(U,\mathscr{C})$ be a fuzzy $\gamma^{\ast}-$covering approximation space, where $U=\{x_{1},x_{2},...,x_{n}\}$, $\mathscr{C}=\{C_{1},C_{2},...,C_{m}\}$, and $X,Y\in\mathscr{F}(U)$. Then\\
$(1)\underline{MR}^{II}_{(\alpha^{\ast},\beta^{\ast})}(U)=U;\overline{MR}^{II}_{(\alpha^{\ast},\beta^{\ast})}(\emptyset)= \emptyset;\\
(2) X\subseteq Y\Rightarrow\overline{MR}^{II}_{(\alpha^{\ast},\beta^{\ast})}(X)\subseteq \overline{MR}^{II}_{(\alpha^{\ast},\beta^{\ast})}(Y);\\
(3) X\subseteq Y\Rightarrow\underline{MR}^{II}_{(\alpha^{\ast},\beta^{\ast})}(X)\subseteq \underline{MR}^{II}_{(\alpha^{\ast},\beta^{\ast})}(Y);\\
(4)\overline{MR}^{II}_{(\alpha^{\ast},\beta^{\ast})}(X)\cup \overline{MR}^{II}_{(\alpha^{\ast},\beta^{\ast})}(X)\subseteq \overline{MR}^{II}_{(\alpha^{\ast},\beta^{\ast})}(X\cup Y);\\
(5) \underline{MR}^{II}_{(\alpha^{\ast},\beta^{\ast})}(X)\cup \underline{MR}^{II}_{(\alpha^{\ast},\beta^{\ast})}(X)\subseteq \underline{MR}^{II}_{(\alpha^{\ast},\beta^{\ast})}(X\cup Y);\\
(6) \overline{MR}^{II}_{(\alpha^{\ast},\beta^{\ast})}(X\cap Y)\subseteq \overline{MR}^{II}_{(\alpha^{\ast},\beta^{\ast})}(X)\cap \overline{MR}^{II}_{(\alpha^{\ast},\beta^{\ast})}(X); \\ (7)\underline{MR}^{II}_{(\alpha^{\ast},\beta^{\ast})}(X\cap Y)\subseteq \underline{MR}^{II}_{(\alpha^{\ast},\beta^{\ast})}(X)\cap \underline{MR}^{II}_{(\alpha^{\ast},\beta^{\ast})}(Y);\\
(8) \alpha^{\ast}_{1} \leq \alpha^{\ast}_{2},\beta^{\ast}_{1} \leq \beta^{\ast}_{2} \Rightarrow\underline{MR}^{II}_{(\alpha^{\ast}_{2},\beta^{\ast}_{2})}(X)\subseteq \underline{MR}^{II}_{(\alpha^{\ast}_{1},\beta^{\ast}_{1})}(Y); \\
(9) \alpha^{\ast}_{1} \leq \alpha^{\ast}_{2},\beta^{\ast}_{1} \leq \beta^{\ast}_{2}  \Rightarrow\overline{MR}^{II}_{(\alpha^{\ast}_{2},\beta^{\ast}_{2})}(X)\subseteq \overline{MR}^{II}_{(\alpha^{\ast}_{1},\beta^{\ast}_{1})}(X).$
\end{theorem}

\noindent\textbf{Proof.}
By Theorem 4.7 and Definition 6.9, the proof is straightforward.
$\Box$

\subsection{Multi-granulation grade approximation operators}

In this section, we present the fuzzy $\gamma^{\ast}-$coverings based multi-granulation grade lower and upper approximation operators for the fuzzy $\gamma^{\ast}-$coverings information system.

\begin{definition}
Let $K_{1}$ and $K_{2}$ be two vectors, where $K_{1}=[k_{11},k_{12},...,k_{1m}]$ and $K_{2}=[k_{21},k_{22},...,k_{2m}]$. It is said that $K_{1}\leq K_{2}$ if we have $k_{1i}\leq k_{2i}$ for any $1\leq i\leq m$.
\end{definition}

In what follows, we first present the fuzzy $\gamma^{\ast}-$coverings based disjunctive
multi-granulation grade lower and upper approximation operators for the fuzzy $\gamma^{\ast}-$coverings information system.

\begin{definition}
Let $(U,\Delta)$ be a fuzzy $\gamma^{\ast}-$coverings information system, and $k_{i}\in R$. Then the fuzzy $\gamma^{\ast}-$coverings based disjunctive multi-granulation grade lower and upper approximations of the fuzzy set $X\in\mathscr{F}(U)$ are defined as follows:
\begin{eqnarray*}
\overline{MR}^{I}_{K}(X)&=&\{x\in U\mid \bigwedge_{1\leq i\leq m}[\Sigma_{y\in U}(X\cap \widetilde{N}_{x}^{\gamma_{i}})(y)>k_{i}]\};\\
\underline{MR}^{I}_{K}(X)&=&\{x\in U\mid \bigwedge_{1\leq i\leq m}[\Sigma_{y\in U}(X^{c}\cap \widetilde{N}_{x}^{\gamma_{i}})(y)\leq k_{i}]\}.
\end{eqnarray*}
\end{definition}

The fuzzy $\gamma^{\ast}-$coverings based disjunctive multi-granulation grade lower and upper approximations operators $\overline{MR}^{I}_{K}(X)$ and $\underline{MR}^{I}_{K}(X)$ of the fuzzy set $X\in\mathscr{F}(U)$ consider the absolute quantitative information, which are generalizations of the disjunctive multi-granulation rough set model proposed by Qian\cite{Qian3}.

\begin{example}
(Continuation from Example 4.4) Taking $k=2$, we have the fuzzy $\gamma^{\ast}-$coverings based disjunctive multi-granulation grade lower and upper approximations of the fuzzy set $X$ as follows:
\begin{eqnarray*}
\overline{GR}_{2}(X)=\{x_{1},x_{2},x_{3},x_{4},x_{5},x_{6},x_{7},x_{8}\} \text{ and }
\underline{GR}_{2}(X)=\{x_{2},x_{3},x_{6},x_{8}\}.
\end{eqnarray*}
\end{example}

We employ an example to illustrate the construction of the fuzzy $\gamma^{\ast}-$coverings based disjunctive multi-granulation probabilistic lower and upper approximations as follows.

\begin{example}(Continuation of Example 6.2) Taking $K=[k_{1},k_{2}]$, where $k_{1}=k_{2}=2$, we have
\begin{eqnarray*}
\underline{MR}^{I}_{K}(X)=\{x_{2},x_{3},x_{6},x_{8}\}\text{ and }
\overline{MR}^{I}_{K}(X)=\{x_{1},x_{2},x_{3},x_{4},x_{5},x_{6},x_{7},x_{8}\}.
\end{eqnarray*}
\end{example}

We employ the following theorem to illustrate the relationship between the fuzzy $\gamma^{\ast}-$coverings based disjunctive multi-granulation grade approximations operators and the fuzzy $\gamma^{\ast}-$coverings based grade approximations operators.

\begin{theorem}
Let $(U,\Delta)$ be a fuzzy $\gamma^{\ast}-$coverings information system,
$0\leq \beta_{i} \leq \alpha_{i} \leq 1, 1\leq i\leq m$, and $X\in\mathscr{F}(U)$. Then we have
\begin{eqnarray*}
\overline{MR}^{I}_{K}(X)&=&\bigcap_{1\leq i\leq m}\overline{GR}_{k_{i}}(X);\\
\underline{MR}^{I}_{K}(X)&=&\bigcap_{1\leq i\leq m}\underline{GR}_{k_{i}}(X).
\end{eqnarray*}
\end{theorem}

\noindent\textbf{Proof.}
The proof is straightforward by Definition 6.15.
$\Box$

\begin{example}(Continuation of Example 6.2) Taking $k_{1}=k_{2}=2,$ we have
\begin{eqnarray*}
\underline{MR}^{I}_{K}(X)&=&\underline{GR}_{k_{1}}(X)\cap \underline{GR}_{k_{2}}(X)\\
&=&\{x_{2},x_{3},x_{6},x_{8}\}\cap \{x_{1},x_{2},x_{3},x_{4},x_{5},x_{6},x_{7},x_{8}\}\\
&=&\{x_{2},x_{3},x_{6},x_{8}\};\\
\overline{MR}^{I}_{K}(X)
&=&\overline{GR}_{k_{1}}(X)\cap \overline{GR}_{k_{2}}(X)\\
&=&\{x_{1},x_{2},x_{3},x_{4},x_{5},x_{6},x_{7},x_{8}\}\cap
\{x_{1},x_{2},x_{3},x_{4},x_{5},x_{6},x_{7},x_{8}\}\\
&=&\{x_{1},x_{2},x_{3},x_{4},x_{5},x_{6},x_{7},x_{8}\}.
\end{eqnarray*}
\end{example}

We study the basic properties of the fuzzy $\gamma^{\ast}-$coverings based disjunctive multi-granulation grade lower and upper approximations operators as follows.

\begin{theorem}
Let $(U,\Delta)$ be a fuzzy $\gamma^{\ast}-$coverings information system,  $K=[k_{1},k_{2},...,k_{m}]$, $K_{1}=[k_{11},k_{12},...,k_{1m}]$, $K_{2}=[k_{21},k_{22},...,k_{2m}]$, and $X,Y\in\mathscr{F}(U)$. Then\\
$(1) \underline{MR}^{I}_{K}(U)=U;\overline{MR}^{I}_{K}(\emptyset)= \emptyset;\\
(2) X\subseteq Y\Rightarrow\overline{MR}^{I}_{K}(X)\subseteq \overline{MR}^{I}_{K}(Y); \\
(3) X\subseteq Y\Rightarrow\underline{MR}^{I}_{K}(X)\subseteq \underline{MR}^{I}_{K}(Y);\\
(4) \overline{MR}^{I}_{K}(X)\cup \overline{MR}^{I}_{K}(X)\subseteq\overline{MR}^{I}_{K}(X\cup Y);\\
(5) \underline{MR}^{I}_{K}(X)\cup \underline{MR}^{I}_{K}(X)\subseteq\underline{MR}^{I}_{K}(X\cup Y);\\
(6) \overline{MR}^{I}_{K}(X\cap Y)\subseteq \overline{MR}^{I}_{K}(X)\cap \overline{MR}^{I}_{K}(Y);\\
(7) \underline{MR}^{I}_{K}(X\cap Y)\subseteq \underline{MR}^{I}_{K}(X)\cap \underline{MR}^{I}_{K}(Y);\\
(8) K_{1} \leq K_{2}\Rightarrow\underline{MR}^{I}_{K_{2}}(X)\subseteq \underline{MR}^{I}_{K_{1}}(X);\\ (9) K_{1} \leq K_{2} \Rightarrow\overline{MR}^{I}_{K_{1}}(X)\subseteq \overline{MR}^{I}_{K_{2}}(X).$
\end{theorem}

\noindent\textbf{Proof.}
By Theorem 4.9 and Definition 6.15, the proof is straightforward.
$\Box$

In what follows, we present the fuzzy $\gamma^{\ast}-$coverings based conjunctive multi-granulation grade lower and upper approximation operators for the fuzzy $\gamma^{\ast}-$covering information system.

\begin{definition}
Let $(U,\Delta)$ be a fuzzy $\gamma^{\ast}-$coverings information system, and $k_{i}\in R$. Then the fuzzy $\gamma^{\ast}-$coverings based conjunctive multi-granulation grade lower and upper approximations of the fuzzy set $X\in\mathscr{F}(U)$ are defined as follows:
\begin{eqnarray*}
\overline{MR}^{II}_{K}(X)&=&\{x\in U\mid \bigvee_{1\leq i\leq m}[\Sigma_{y\in U}(X\cap \widetilde{N}_{x}^{\gamma_{i}})(y)>k_{i}]\};\\
\underline{MR}^{II}_{K}(X)&=&\{x\in U\mid \bigvee_{1\leq i\leq m}[\Sigma_{y\in U}(X^{c}\cap \widetilde{N}_{x}^{\gamma_{i}})(y)\leq k_{i}]\}.
\end{eqnarray*}
\end{definition}

The fuzzy $\gamma^{\ast}-$coverings based conjunctive multi-granulation grade lower and upper approximations operators $\overline{MR}^{II}_{K}(X)$ and $\underline{MR}^{II}_{K}(X)$ of the fuzzy set $X\in\mathscr{F}(U)$ consider the absolute quantitative information, which are generalizations of multi-granulation rough set model proposed by Qian\cite{Yao}.

We employ an example to illustrate the construction of the fuzzy $\gamma^{\ast}-$coverings based conjunctive multi-granulation grade lower and upper approximations of sets as follows.

\begin{example}(Continuation of Example 5.2) Taking $X=\frac{0.6}{x_{1}}+\frac{0.5}{x_{2}}+\frac{0.7}{x_{3}}
+\frac{0.8}{x_{4}}+\frac{0.5}{x_{5}}
+\frac{0.6}{x_{6}}+\frac{0}{x_{7}}+\frac{0.2}{x_{8}}$, $K=[k_{1},k_{2}]$, where $k_{1}=k_{2}=2$. Then we have
\begin{eqnarray*}
\underline{MR}^{II}_{K}(X)=\{x_{1},x_{2},x_{3},x_{4},x_{5},x_{6},x_{7},x_{8}\}\text{ and }
\overline{MR}^{II}_{K}(X)=\{x_{1},x_{2},x_{3},x_{4},x_{5},x_{6},x_{7},x_{8}\}.
\end{eqnarray*}
\end{example}

We employ the following theorem to illustrate the relationship between the fuzzy $\gamma^{\ast}-$coverings based conjunctive multi-granulation grade approximations operators and the fuzzy $\gamma^{\ast}-$coverings based grade approximations operators.

\begin{theorem}
Let $(U,\Delta)$ be a fuzzy $\gamma^{\ast}-$coverings information system,
$0\leq \beta_{i} \leq \alpha_{i} \leq 1, 1\leq i\leq m$, and $X\in\mathscr{F}(U)$. Then we have
\begin{eqnarray*}
\overline{MR}^{II}_{K}(X)&=&\bigcup_{1\leq i\leq m}\overline{GR}_{k_{i}}(X);\\
\underline{MR}^{II}_{K}(X)&=&\bigcup_{1\leq i\leq m}\underline{GR}_{k_{i}}(X).
\end{eqnarray*}
\end{theorem}

\noindent\textbf{Proof.}
The proof is straightforward by Definition 6.21.
$\Box$

\begin{example}(Continuation of Example 6.4) Taking $k_{1}=k_{2}=2,$ we have
\begin{eqnarray*}
\underline{MR}^{II}_{K}(X)&=&\underline{GR}_{k_{1}}(X)\cup \underline{GR}_{k_{2}}(X)\\
&=&\{x_{2},x_{3},x_{6},x_{8}\}\cup \{x_{1},x_{2},x_{3},x_{4},x_{5},x_{6},x_{7},x_{8}\}\\
&=&\{x_{1},x_{2},x_{3},x_{4},x_{5},x_{6},x_{7},x_{8}\};\\
\overline{MR}^{II}_{K}(X)
&=&\overline{GR}_{k_{1}}(X)\cup \overline{GR}_{k_{2}}(X)\\
&=&\{x_{1},x_{2},x_{3},x_{4},x_{5},x_{6},x_{7},x_{8}\}\cup
\{x_{1},x_{2},x_{3},x_{4},x_{5},x_{6},x_{7},x_{8}\}\\
&=&\{x_{1},x_{2},x_{3},x_{4},x_{5},x_{6},x_{7},x_{8}\}.
\end{eqnarray*}
\end{example}

We show the basic properties of the fuzzy $\gamma^{\ast}-$coverings based conjunctive multi-granulation grade lower and upper approximations operators as follows.

\begin{theorem}
Let $(U,\Delta)$ be a fuzzy $\gamma^{\ast}-$coverings information system,  $K=[k_{1},k_{2},...,k_{m}]$, $K_{1}=[k_{11},k_{12},...,k_{1m}]$, $K_{2}=[k_{21},k_{22},...,k_{2m}]$, and $X,Y\in\mathscr{F}(U)$. Then\\
$(1) \underline{MR}^{II}_{K}(U)=U;\overline{MR}^{II}_{K}(\emptyset)= \emptyset;\\
(2) X\subseteq Y\Rightarrow\overline{MR}^{II}_{K}(X)\subseteq \overline{MR}^{II}_{K}(Y); \\
(3) X\subseteq Y\Rightarrow\underline{MR}^{II}_{K}(X)\subseteq \underline{MR}^{II}_{K}(Y);\\
(4) \overline{MR}^{II}_{K}(X)\cup \overline{MR}^{II}_{K}(X)\subseteq\overline{MR}^{II}_{K}(X\cup Y);\\
(5) \underline{MR}^{II}_{K}(X)\cup \underline{MR}^{II}_{K}(X)\subseteq\underline{MR}^{II}_{K}(X\cup Y);\\
(6) \overline{MR}^{II}_{K}(X\cap Y)\subseteq \overline{MR}^{II}_{K}(X)\cap \overline{MR}^{II}_{K}(Y);\\
(7) \underline{MR}^{II}_{K}(X\cap Y)\subseteq \underline{MR}^{II}_{K}(X)\cap \underline{MR}^{II}_{K}(Y);\\
(8) K_{1} \leq K_{2}\Rightarrow\underline{MR}^{II}_{K_{2}}(X)\subseteq \underline{MR}^{II}_{K_{1}}(X);\\ (9) K_{1} \leq K_{2} \Rightarrow\overline{MR}^{II}_{K_{1}}(X)\subseteq \overline{MR}^{II}_{K_{2}}(X).$
\end{theorem}

\noindent\textbf{Proof.}
By Theorem 4.9 and Definition 6.21, the proof is straightforward.
$\Box$

\section{Multi-granulation double-quantitative approximation operators}

In this section, we provide the fuzzy $\gamma^{\ast}-$coverings based multi-granulation double-quantitative lower and upper approximation operators in the fuzzy $\gamma-$coverings information system.

\begin{definition}
Let $(U,\Delta)$ be a fuzzy $\gamma^{\ast}-$coverings information system, $0\leq \beta_{i} \leq \alpha_{i} \leq 1,$ and $ k_{i}\in R $. Then the fuzzy $\gamma^{\ast}-$coverings based disjunctive multi-granulation double-quantitative lower and upper approximations of the fuzzy set $X\in \mathscr{F}(U)$ are defined as follows:
\begin{eqnarray*}
\overline{MR}_{(\alpha^{\ast},\beta^{\ast})\wedge K}^{I}(X)&=&\{x\in U\mid \bigwedge_{1\leq i\leq m}[P(X|\widetilde{N}_{x}^{\gamma_{i}})\geq\beta^{\ast}_{i} \wedge \Sigma_{y\in U}(X\cap \widetilde{N}_{x}^{\gamma_{i}})(y)>k_{i}]\};\\
\underline{MR}_{(\alpha^{\ast},\beta^{\ast})\wedge K}^{I}(X)&=&\{x\in U\mid \bigwedge_{1\leq i\leq m}[P(X|\widetilde{N}_{x}^{\gamma_{i}})\geq\alpha^{\ast}_{i} \wedge \Sigma_{y\in U}(X^{c}\cap \widetilde{N}_{x}^{\gamma_{i}})(y)\leq k_{i}]\}.
\end{eqnarray*}
\end{definition}

The fuzzy $\gamma^{\ast}-$coverings based disjunctive multi-granulation double-quantitative lower and upper approximation operators $\overline{MR}_{(\alpha^{\ast},\beta^{\ast})\wedge K}^{I}(X)$ and $\underline{MR}_{(\alpha^{\ast},\beta^{\ast})\wedge K}^{I}(X)$ of the fuzzy set $X\in \mathscr{F}(U)$ consider the absolute and relative quantitative information, which are generalizations of disjunctive double quantitative rough set model proposed by Xu\cite{Xu1}.

\begin{theorem}
Let $(U,\Delta)$ be a fuzzy $\gamma^{\ast}-$coverings information system, $0\leq \beta_{i} \leq \alpha_{i} \leq 1,$ and $ k_{i}\in R $. Then the fuzzy $\gamma^{\ast}-$coverings based disjunctive multi-granulation double-quantitative lower and upper approximations of the fuzzy set $X\in \mathscr{F}(U)$ are defined as follows:
\begin{eqnarray*}
\overline{MR}_{(\alpha^{\ast},\beta^{\ast})\wedge K}^{I}(X)&=&\overline{MR}_{(\alpha^{\ast},\beta^{\ast})}^{I}(X) \cap \overline{MR}_{K}^{I}(X);\\
\underline{MR}_{(\alpha^{\ast},\beta^{\ast})\wedge K}^{I}(X)&=&\underline{MR}_{(\alpha^{\ast},\beta^{\ast})}^{I}(X)\cap \underline{MR}_{ K}^{I}(X).
\end{eqnarray*}
\end{theorem}

\noindent\textbf{Proof.}
The proof is straightforward by Definition 7.1.
$\Box$

\begin{example}(Continuation of Example 6.4) Taking $\alpha_{1}=\alpha_{2}=0.75,$ and $\beta_{1}=\beta_{2}=0.25,k_{1}=k_{2}=1$ we have
\begin{eqnarray*}
\overline{MR}_{(\alpha^{\ast},\beta^{\ast})\wedge K}^{I}(X)&=&\overline{MR}_{(\alpha^{\ast},\beta^{\ast})}^{I}(X) \cap \overline{MR}_{K}^{I}(X)\\
&=&\{x_{3},x_{6}\}\cap \{x_{2},x_{3},x_{6},x_{8}\}\\
&=&\{x_{3},x_{6}\};\\
\underline{MR}_{(\alpha^{\ast},\beta^{\ast})\wedge K}^{I}(X)&=&\underline{MR}_{(\alpha^{\ast},\beta^{\ast})}^{I}(X)\cap \underline{MR}_{ K}^{I}(X)\\
&=&\{x_{1},x_{2},x_{3},x_{4},x_{5},x_{6},x_{7},x_{8}\}\cap \{x_{1},x_{2},x_{3},x_{4},x_{5},x_{6},x_{7},x_{8}\}\\
&=&\{x_{1},x_{2},x_{3},x_{4},x_{5},x_{6},x_{7},x_{8}\}.
\end{eqnarray*}
\end{example}

We show the basic properties of the fuzzy $\gamma^{\ast}-$coverings based disjunctive multi-granulation double-quantitative lower and upper approximation operators as follows.

\begin{theorem}
Let $(U,\Delta)$ be a fuzzy $\gamma^{\ast}-$coverings information system, and $X,Y\in\mathscr{F}(U)$. Then\\
$(1)\underline{MR}_{(\alpha^{\ast},\beta^{\ast})\wedge K}^{I}(U)=U;\overline{MR}_{(\alpha^{\ast},\beta^{\ast})\wedge K}^{I}(\emptyset)= \emptyset;\\
(2) X\subseteq Y\Rightarrow\overline{MR}_{(\alpha^{\ast},\beta^{\ast})\wedge K}^{I}(X)\subseteq \overline{MR}_{(\alpha^{\ast},\beta^{\ast})\wedge K}^{I}(Y);\\
(3) X\subseteq Y\Rightarrow\underline{MR}_{(\alpha^{\ast},\beta^{\ast})\wedge k}^{I}(X)\subseteq \underline{MR}_{(\alpha^{\ast},\beta^{\ast})\wedge K}^{I}(Y);\\
(4) \overline{MR}_{(\alpha^{\ast},\beta^{\ast})\wedge K}^{I}(X)\cup \overline{MR}_{(\alpha^{\ast},\beta^{\ast})\wedge K}^{I}(Y)\subseteq \overline{MR}_{(\alpha^{\ast},\beta^{\ast})\wedge K}^{I}(X\cup Y);\\
(5) \underline{MR}_{(\alpha^{\ast},\beta^{\ast})\wedge K}^{I}(X)\cup \underline{MR}_{(\alpha^{\ast},\beta^{\ast})\wedge K}^{I}(Y)\subseteq \underline{MR}_{(\alpha^{\ast},\beta^{\ast})\wedge K}^{I}(X\cup Y);\\
(6) \overline{MR}_{(\alpha^{\ast},\beta^{\ast})\wedge K}^{I}(X\cap Y)\subseteq \overline{MR}_{(\alpha^{\ast},\beta^{\ast})\wedge K}^{I}(X)\cap \overline{MR}_{(\alpha^{\ast},\beta^{\ast})\wedge K}^{I}(Y);\\
(7) \underline{MR}_{(\alpha^{\ast},\beta^{\ast})\wedge K}^{I}(X\cap Y)\subseteq \underline{MR}_{(\alpha^{\ast},\beta^{\ast})\wedge K}^{I}(X)\cap \underline{MR}_{(\alpha^{\ast},\beta^{\ast})\wedge K}^{I}(Y);\\
(8) \alpha^{\ast}_{1} \leq \alpha^{\ast}_{2}, \beta^{\ast}_{1} \leq \beta^{\ast}_{2}, K_{1} \leq K_{2}\Rightarrow\underline{MR}_{(\alpha^{\ast}_{1},\beta^{\ast}_{1})\wedge K_{1}}^{I}(X)\subseteq \underline{MR}_{(\alpha^{\ast}_{2},\beta^{\ast}_{2})\wedge k_{2}}^{I}(X);\\
(9) \alpha^{\ast}_{1} \leq \alpha^{\ast}_{2}, \beta^{\ast}_{1} \leq \beta^{\ast}_{2}, K_{1} \leq K_{2} \Rightarrow\overline{MR}_{(\alpha^{\ast}_{1},\beta^{\ast}_{1})\wedge K_{1}}^{I}(X)\subseteq \overline{MR}_{(\alpha^{\ast}_{2},\beta^{\ast}_{2})\wedge K_{2}}^{I}(X).$
\end{theorem}

\noindent\textbf{Proof.}
By Theorem 4.14, the proof is straightforward.$\Box$

\begin{definition}
Let $(U,\Delta)$ be a fuzzy $\gamma^{\ast}-$coverings information system, $0\leq \beta_{i} \leq \alpha_{i} \leq 1,$ and $ k_{i}\in R $. Then the fuzzy $\gamma^{\ast}-$coverings based conjunctive multi-granulation double-quantitative lower and upper approximations of the fuzzy set $X\in \mathscr{F}(U)$ are defined as follows:
\begin{eqnarray*}
\overline{MR}_{(\alpha^{\ast},\beta^{\ast})\vee K}^{II}(X)&=&\{x\in U\mid \bigvee_{1\leq i\leq m}[P(X|\widetilde{N}_{x}^{\gamma_{i}})\geq\beta^{\ast}_{i} \vee \Sigma_{y\in U}(X\cap \widetilde{N}_{x}^{\gamma_{i}})(y)>k_{i}]\};\\
\underline{MR}_{(\alpha^{\ast},\beta^{\ast})\vee K}^{II}(X)&=&\{x\in U\mid \bigvee_{1\leq i\leq m}[P(X|\widetilde{N}_{x}^{\gamma_{i}})\geq\alpha^{\ast}_{i} \vee \Sigma_{y\in U}(X^{c}\cap \widetilde{N}_{x}^{\gamma_{i}})(y)\leq k_{i}]\}.
\end{eqnarray*}
\end{definition}

The fuzzy $\gamma^{\ast}-$coverings based conjunctive multi-granulation double-quantitative lower and upper approximation operators $\overline{MR}_{(\alpha^{\ast},\beta^{\ast})\vee K}^{II}(X)$ and $\underline{MR}_{(\alpha^{\ast},\beta^{\ast})\vee K}^{II}(X)$ of the fuzzy set $X\in \mathscr{F}(U)$ consider the absolute and relative quantitative information, which are generalizations of conjunctive double quantitative rough set model proposed by Xu\cite{Xu1}.

\begin{theorem}
Let $(U,\Delta)$ be a fuzzy $\gamma^{\ast}-$coverings information system, $0\leq \beta_{i} \leq \alpha_{i} \leq 1,$ and $ k_{i}\in R $. Then the fuzzy $\gamma^{\ast}-$coverings based disjunctive multi-granulation double-quantitative lower and upper approximations of the fuzzy set $X\in \mathscr{F}(U)$ are defined as follows:
\begin{eqnarray*}
\overline{MR}_{(\alpha^{\ast},\beta^{\ast})\vee K}^{II}(X)&=&\overline{MR}_{(\alpha^{\ast},\beta^{\ast})}^{II}(X) \cup \overline{MR}_{K}^{II}(X);\\
\underline{MR}_{(\alpha^{\ast},\beta^{\ast})\vee K}^{II}(X)&=&\underline{MR}_{(\alpha^{\ast},\beta^{\ast})}^{II}(X)\cup \underline{MR}_{ K}^{II}(X).
\end{eqnarray*}
\end{theorem}

\noindent\textbf{Proof.}
The proof is straightforward by Definition 7.5.
$\Box$

\begin{example}(Continuation of Example 6.4) Taking $\alpha_{1}=\alpha_{2}=0.75,$ and $\beta_{1}=\beta_{2}=0.25,k_{1}=k_{2}=1$ we have
\begin{eqnarray*}
\overline{MR}_{(\alpha^{\ast},\beta^{\ast})\vee K}^{II}(X)&=&\overline{MR}_{(\alpha^{\ast},\beta^{\ast})}^{II}(X) \cup \overline{MR}_{K}^{II}(X)\\
&=&\{x_{3},x_{6}\}\cup \{x_{2},x_{3},x_{6},x_{8}\}\\
&=&\{x_{2},x_{3},x_{6},x_{8}\};\\
\underline{MR}_{(\alpha^{\ast},\beta^{\ast})\vee K}^{II}(X)&=&\underline{MR}_{(\alpha^{\ast},\beta^{\ast})}^{II}(X)\cup \underline{MR}_{ K}^{II}(X)\\
&=&\{x_{1},x_{2},x_{3},x_{4},x_{5},x_{6},x_{7},x_{8}\}\cup \{x_{1},x_{2},x_{3},x_{4},x_{5},x_{6},x_{7},x_{8}\}\\
&=&\{x_{1},x_{2},x_{3},x_{4},x_{5},x_{6},x_{7},x_{8}\}.
\end{eqnarray*}
\end{example}

We present the basic properties of the fuzzy $\gamma^{\ast}-$coverings based conjunctive multi-granulation double-quantitative lower and upper approximation operators as follows.

\begin{theorem}
Let $(U,\Delta)$ be a fuzzy $\gamma^{\ast}-$covering information system, and $X,Y\in\mathscr{F}(U)$. Then\\
$(1)\underline{MR}_{(\alpha^{\ast},\beta^{\ast})\vee K}^{II}(U)=U;\overline{MR}_{(\alpha^{\ast},\beta^{\ast})\vee K}^{II}(\emptyset)= \emptyset;\\
(2) X\subseteq Y\Rightarrow\overline{MR}_{(\alpha^{\ast},\beta^{\ast})\vee K}^{II}(X)\subseteq \overline{MR}_{(\alpha^{\ast},\beta^{\ast})\vee K}^{I}(Y);\\
(3) X\subseteq Y\Rightarrow\underline{MR}_{(\alpha^{\ast},\beta^{\ast})\vee k}^{II}(X)\subseteq \underline{MR}_{(\alpha^{\ast},\beta^{\ast})\vee K}^{II}(Y);\\
(4) \overline{MR}_{(\alpha^{\ast},\beta^{\ast})\vee K}^{II}(X)\cup \overline{MR}_{(\alpha^{\ast},\beta^{\ast})\vee K}^{II}(Y)\subseteq \overline{MR}_{(\alpha^{\ast},\beta^{\ast})\vee K}^{II}(X\cup Y);\\
(5) \underline{MR}_{(\alpha^{\ast},\beta^{\ast})\vee K}^{II}(X)\cup \underline{MR}_{(\alpha^{\ast},\beta^{\ast})\vee K}^{II}(Y)\subseteq \underline{MR}_{(\alpha^{\ast},\beta^{\ast})\vee K}^{II}(X\cup Y);\\
(6) \overline{MR}_{(\alpha^{\ast},\beta^{\ast})\vee K}^{II}(X\cap Y)\subseteq \overline{MR}_{(\alpha^{\ast},\beta^{\ast})\vee K}^{II}(X)\cap \overline{MR}_{(\alpha^{\ast},\beta^{\ast})\vee K}^{II}(Y);\\
(7) \underline{MR}_{(\alpha^{\ast},\beta^{\ast})\vee K}^{II}(X\cap Y)\subseteq \underline{MR}_{(\alpha^{\ast},\beta^{\ast})\vee K}^{II}(X)\cap \underline{MR}_{(\alpha^{\ast},\beta^{\ast})\vee K}^{II}(Y);\\
(8) \alpha^{\ast}_{1} \leq \alpha^{\ast}_{2}, \beta^{\ast}_{1} \leq \beta^{\ast}_{2}, K_{1} \leq K_{2}\Rightarrow\underline{MR}_{(\alpha^{\ast}_{1},\beta^{\ast}_{1})\vee K_{1}}^{II}(X)\subseteq \underline{MR}_{(\alpha^{\ast}_{2},\beta^{\ast}_{2})\vee k_{2}}^{II}(X);\\
(9) \alpha^{\ast}_{1} \leq \alpha^{\ast}_{2}, \beta^{\ast}_{1} \leq \beta^{\ast}_{2}, K_{1} \leq K_{2} \Rightarrow\overline{MR}_{(\alpha^{\ast}_{1},\beta^{\ast}_{1})\vee K_{1}}^{II}(X)\subseteq \overline{MR}_{(\alpha^{\ast}_{2},\beta^{\ast}_{2})\vee K_{2}}^{I}(X).$
\end{theorem}

\noindent\textbf{Proof.}
By Theorem 4.17, the proof is straightforward.
$\Box$

\section{Conclusions}

In this paper, we have presented the fuzzy $\gamma-$covering based probabilistic and grade lower and upper approximation operators. Second, we have provided the fuzzy $\gamma-$covering based double-quantitative lower and upper approximation operators. Third, we have proposed the fuzzy $\gamma^{\ast}-$coverings based multi-granulation probabilistic and grade lower and upper approximation operators. Fourth, we have presented the fuzzy $\gamma^{\ast}-$coverings based multi-granulation double-quantitative lower and upper approximation operators. Finally, we have employed several examples to illustrate how to construct the lower and upper approximations of fuzzy sets with the relative and absolute information.

There are a lot of fuzzy covering information systems in practical situations, we should further study the fuzzy covering based lower and upper approximation operators and knowledge discovery of fuzzy covering information systems, so as to build the bridge between the fuzzy covering rough set theory and other rough set models in the future.

\section*{ Acknowledgments}

We would like to thank the anonymous reviewers very much for their
professional comments and valuable suggestions. This work is
supported by the National Natural Science Foundation of China (NO.
61673301,61603063,11526039), Doctoral Fund of Ministry of Education of China(No.201300721004), China Postdoctoral Science Foundation(NO.2013M542558,2015M580353), the Scientific
Research Fund of Hunan Provincial Education Department(No.15B004).

\end{document}